**Transferability and operational reliability of a Prithvi crop classification foundation model under phenological and geographic shift across three continents**

Venkatesh Kolluru[1,*], Rajat Shinde[1], Abdelhak Marouane[1], Caden Helbling[1]; Deepak Shah[1], Othneil Drew[2], Srinivas Kolluru[1], Iksha Gurung[1], Manil Maskey[3], Rahul Ramachandran[3]

[1]Earth System Science Center, University of Alabama in Huntsville, Huntsville, AL 35899, USA
[2]Space and Earth Science Data Analysis,
[3]NASA Marshall Space Flight Center

[*]Corresponding author
Venkatesh Kolluru
Earth System Science Center, University of Alabama in Huntsville, Huntsville, AL 35899, USA
Email: venkatesh.kolluru@uah.edu

## Abstract

Fine-tuned geospatial foundation models (GeoFMs) pretrained on large satellite archives have been shown to improve crop classification accuracy and geographic transferability. However, their operational performance beyond the training distribution remains poorly characterized. We evaluated the out-of-distribution performance of a widely adopted GeoFM [Prithvi-EO-2.0, a 600-million-parameter temporal Vision Transformer] across 37 events in 12 countries on three continents and validated against regional reference products (CDL, AAFC, EUCROPMAP, and MapBiomas). Results indicated that the mean overall accuracy (OA) declined from 0.65 in the United States to 0.40 in Europe. Beyond accuracy metrics, we assessed five key aspects of model performance: whether model confidence indicates signal failure, sensitivity to observation windows, the effect of coarsening class schemes, and robustness to both band loss and cloud- and shadow-contamination. Accuracy collapsed when the observation window misaligned with local crop phenology, while deterministic confidence remained high. Expected calibration error increased for seven of eight paired events, and 12-51% of each affected scene was confidently mislabeled at near-zero precision. Monte Carlo dropout entropy registered the shift in all eight, indicating that much of the apparent cross-continent decline reflected phenological misalignment rather than spatial transfer. Two adjustments recovered accuracy without retraining. Consolidating 13 classes into 10, based on the model's dominant confusions, raised OA at every event by a mean of 8.4 percentage points and macro-F1 by 0.070. Compressing the window toward near-real-time use preserved accuracy across a 45- to 90-day plateau, peaking near 75 days, though arms tighter than 30 days fell about 0.11 below that plateau. Prithvi-EO-2.0 exceeded Prithvi-EO-1.0 under an identical protocol, and the red band was the most damaging single loss across all regions. Fine-tuned crop GeoFMs therefore transfer usefully only where observation windows match local growing seasons, a condition that deterministic confidence cannot verify. We translate these findings into operational guidance for the reliable deployment of the released model.

## 1. Introduction

Timely and spatially explicit crop-type information is crucial for agricultural monitoring, yield forecasting, water allocation, and food-security assessment (Zhang et al., 2025). Though this information is needed globally, the labeled reference data used to produce it are organized regionally (Kolluru et al., 2026b). This mismatch is primarily due to the national and continental mapping programs being built around local cropping systems, growing calendars, and crop classification systems. The United States Cropland Data Layer (CDL), the Agriculture and Agri-Food Canada Annual Crop Inventory (AAFC), and the pan-European EUCROPMAP each map crops wall-to-wall within a single jurisdiction, and recent literature counts more than sixty such regional products with largely incompatible legends (Zhang et al., 2025). Because each product is calibrated to its own region and season, a crop map accurate in one jurisdiction cannot be assumed to hold in another.

Crop-type models degrade when transferred beyond their training region, a limitation that predates geospatial foundation models (GeoFMs). Supervised classifiers of satellite image time series would separate spectrally similar crops based on phenological trajectories and achieve overall accuracies above 80% within the training region when trained on local in situ samples (Jain et al., 2024). However, their accuracy declines in new regions because cropping calendars, class priors, and sensor conditions differ (Sun et al., 2020; Long et al., 2025). The domain-adaptation literature quantified this loss and developed explicit corrections using several techniques, including spatially generalizable architectures, explicit temporal-shift and climate-indicator alignment, label-free transfer, and dedicated cross-region and few-shot benchmarks (Wang et al., 2019; Xu et al., 2020; Tseng et al., 2021; Nyborg et al., 2022; Koukos et al., 2024; Ma et al., 2024; Reuss et al., 2025; Wang et al., 2025). These approaches helped recover 6-20 points in overall accuracy or macro-F1 across regions without any source-region data (Long et al., 2025; Mohammadi et al., 2024; Wang et al., 2022, 2023). Corrections of this kind, however, typically require roughly three thousand labeled samples per class in the target region to restore accuracy (Long et al., 2025).

GeoFMs were proposed to reduce this dependence on region-specific training data by pretraining on large unlabeled Earth observation (EO) archives via self-supervision. The premise is that such models learn general-purpose representations that a lightweight decoder can adapt to many downstream tasks with limited labeled data (Mai et al., 2023). Several such models have since been developed for EO, including Prithvi (Jakubik et al., 2023), SatlasPretrain (Bastani et al., 2023), and TerraMind (Jakubik et al., 2025), etc. Prithvi GeoFMs, developed by NASA and IBM, are among the most widely used because they were pretrained on Harmonized Landsat and Sentinel-2 (HLS) imagery, a public 30 m dataset well suited to crop monitoring, and are explicitly multi-temporal, which aligns with the phenological signals on which crop classification relies. Prithvi-EO-1.0 introduced masked autoencoder pretraining on HLS imagery (Jakubik et al., 2023), and Prithvi-EO-2.0 increased backbone capacity and added temporal and location metadata conditioning (Szwarcman et al., 2025).

Two distinct pathways for transferring these models to new regions have since been tested. The first assumes that pretraining alone produces features general enough for zero-shot or frozen-embedding transfer. However, evidence now suggests that this pathway fails to map crops across continents (Sachdeva et al., 2024; Chang et al., 2025). AlphaEarth embeddings and frozen Prithvi-EO-1.0 embeddings that achieve $R^2$ near 0.80 for United States corn and soybean collapse under leave-one-country-out transfer to Argentina or sub-Saharan Africa, offering no advantage over simple spectral baselines (Ma et al., 2026; Adjei, 2026). Systematic benchmarks further find that GeoFMs do not consistently outperform strong supervised baselines (Huang et al., 2026) and that inconsistent protocols make it difficult to establish the state of the art, motivating protocol-explicit and variance-aware evaluation (Marsocci et al., 2024; Corley et al., 2026). The second pathway, fine-tuning on target-region data or adaptation, restores accuracy (Li et al., 2025), as shown by Prithvi-EO-2.0, fine-tuned for smallholder crop-type and crop-health mapping in India (Tomotaki-Dawoud et al., 2025), canola-yield regression on the Canadian Prairies (Nejadshamsi et al., 2026), and the Presto fine-tuned on globally distributed WorldCereal labels (Butsko et al., 2025). However, operational users most often face a third scenario in which they apply a released, already fine-tuned checkpoint to a new region and season without further adaptation. Because the accuracy of such checkpoints depends on the fine-tuning distribution, their behavior in out-of-distribution deployment has not been characterized at the map level for any released model. Consequently, we do not know how far a released checkpoint degrades, which deployment conditions drive the degradation, or whether the model flags its own errors when they occur. Therefore, the reliability of such deployments remains an open question.

Prior studies typically report only a single accuracy number or an out-of-distribution accuracy figure, which conflates geographic shift with the timing of the observation window relative to local phenology (Nyborg et al., 2022; Stehman and Foody, 2019) and the observation conditions of the input imagery. Gonzalez-Calabuig et al. (2026) also demonstrated that single-model probabilistic confidence is an unreliable indicator of failure and that ensemble or epistemic uncertainty is necessary to flag out-of-regime inputs. No prior study has tested whether these findings on confidence reliability extend to a crop-classification model released for operational use and deployed at a map scale across continents. Therefore, we followed established remote-sensing practice by isolating these factors experimentally and reporting class-level estimates and their uncertainties rather than a single summary statistic (Olofsson et al., 2014; Kolluru et al., 2024). We addressed this gap by evaluating the released Prithvi-EO-2.0-600M-TL crop-classification model across 37 out-of-distribution (OOD) events spanning 12 countries and three continents. We validate each event against the reference product for its region, namely, the CDL in the United States, the AAFC in Canada, EUCROPMAP in Europe, and MapBiomas in South America (Souza et al., 2020), with all inputs drawn from HLS surface reflectance (Claverie et al., 2018). We also benchmark the model against its Prithvi-EO-1.0-100M predecessor under the identical protocol. This comparison isolates the effect of the backbone upgrade and tests whether the in-distribution performance gains of the larger V2 model propagate to OOD deployment. Furthermore, this evaluation is neither a zero-shot experiment nor a target-region adaptation study. It is a deployment test of a released fine-tuned model applied without further modification, which is the situation an operational user faces when applying a published checkpoint to a new region (Kolluru et al., 2026a).

The evaluation is organized around six research questions that address the shortcomings and gaps identified above, as well as the drivers of OOD performance. RQ1 asks how well the released model generalizes to OOD events across multiple continents, and how between-event accuracy varies along the distance gradient from the training distribution. RQ2 asks whether predicted confidence fails under changes to the three temporal frames or whether incorrect predictions retain high confidence. RQ3 asks how far the three-frame window can be compressed toward a near-real-time configuration before accuracy declines. RQ4 asks how much accuracy a coarser class scheme recovers and whether the gain extends to class-balanced metrics. RQ5a asks whether the V1-to-V2 upgrade that improves in-distribution benchmarks also helps under OOD conditions. RQ5b asks how robust the model is to band corruption and to cloud and shadow contamination, and which spectral bands matter most for crop discrimination.

Questions RQ1, RQ4, RQ5a, and RQ5b correspond to the sensitivities that recent syntheses identify as governing whether a foundation model's generality survives deployment. These sensitivities include label quality, sensor heterogeneity, and pretraining data (Huang et al., 2026). They also include the domain-shift conditions cataloged in transfer-learning reviews (Ma et al., 2024; Long et al., 2025). Questions RQ2 and RQ3 extend beyond these factors. RQ2 addresses the reliability of the model's reported confidence under temporal misalignment, and RQ3 addresses the position of the observation window within the growing season. Neither dimension has been measured at the map scale for crop foundation models (Nyborg et al., 2022; Gonzalez-Calabuig et al., 2026). Furthermore, reporting all six questions under a single protocol answers recent calls for protocol-explicit and variance-aware OOD evaluation (Marsocci et al., 2024; Corley et al., 2026) and complements the curated downstream task benchmarks now being assembled for foundation models in other Earth science domains (Shinde et al., 2026).

This study makes five empirical contributions that answer the questions above. First, it provides the first operational, map-level, multi-continent OOD evaluation of the released fine-tuned Prithvi-EO-2.0 crop model in its intended configuration. Second, it isolates the effect of phenological-window misalignment from geographic transfer by holding location fixed while varying only the observation window. Third, it reports a within-year rolling-window analysis of temporal stability and a near-real-time feasibility analysis for a crop foundation model. Fourth, it quantifies a class-scheme reorganization within a single, consistent accounting framework and delivers a head-to-head comparison of Prithvi V1 and V2 for multi-class crop mapping beyond the training distribution, under an identical protocol. Fifth, it characterizes the robustness of band corruption in a Prithvi crop model. Together, these analyses translate the sensitivities described only qualitatively in recent reviews into reproducible map-level measurements for a model already in operational use (Huang et al., 2026; Marsocci et al., 2024).

## 2. Data and methods

### 2.1 Overview

This study evaluates whether the fine-tuned Prithvi-EO-2.0-600M-TL crop-classification model is generalizable and remains accurate when applied to new geographies, seasons, sensor conditions, and reference nomenclatures without further adaptation. The study area and the evaluation workflow are summarized in Figures 1 and 2. The evaluation workflow comprises: (a) assembly of 37 globally distributed OOD events; (b) acquisition and harmonization of HLS surface reflectance time series; (c) inference with the released model and its Prithvi-EO-1.0-100M predecessor under an identical protocol; (d) reclassification of each regional reference product into a shared label space and pixel-wise validation under an exclude-from-denominator convention; (e) computation of a multi-metric accuracy and calibration suite; and (f) six controlled experiments that stress the deployed model along the operational axes defined by the research questions (Fig. 2). The model weights are frozen throughout so that only the deployment conditions vary between events and experiments.

### 2.2 The released model and its configuration

The primary model is the released fine-tuned Prithvi-EO-2.0-600M-TL crop-classification model (Szwarcman et al., 2025). It couples a ~600M-parameter Vision Transformer (ViT) encoder, pretrained by masked autoencoding on a global HLS archive, with a lightweight decoder fine-tuned for multi-temporal crop segmentation on the HuggingFace multi-temporal crop-classification dataset (approximately 5,000 United States HLS samples). The task is 13-class semantic segmentation covering natural vegetation, forest, corn, soybeans, wetlands, developed/barren, open water, winter wheat, alfalfa, fallow/idle cropland, cotton, sorghum, and other. The input is a three-frame (T=3) time series of six HLS surface reflectance bands (blue, green, red, narrow near-infrared, shortwave infrared 1, shortwave infrared 2) at 30 m resolution, arranged into 224×224 patches.

We ran the model using the TerraTorch inference toolkit (Gomes et al., 2025) with the backbone frozen and the decoder fine-tuned with a weighted cross-entropy objective. The checkpoint uses the temporal- and location-aware (TL) backbone variant, where pretraining randomly dropped geolocation and acquisition-time inputs, enabling the model to operate without them (Szwarcman et al., 2025). The coordinate encoding was left empty when the released crop model was fine-tuned, and we ran inference in the same configuration, supplying no latitude, longitude, or acquisition-time metadata. For the backbone-generation comparison (RQ5a), we also run the released Prithvi-EO-1.0-100M crop model (Jakubik et al., 2023) under the same protocol, including the same input chips, 13 classes, reference validation, and scoring.

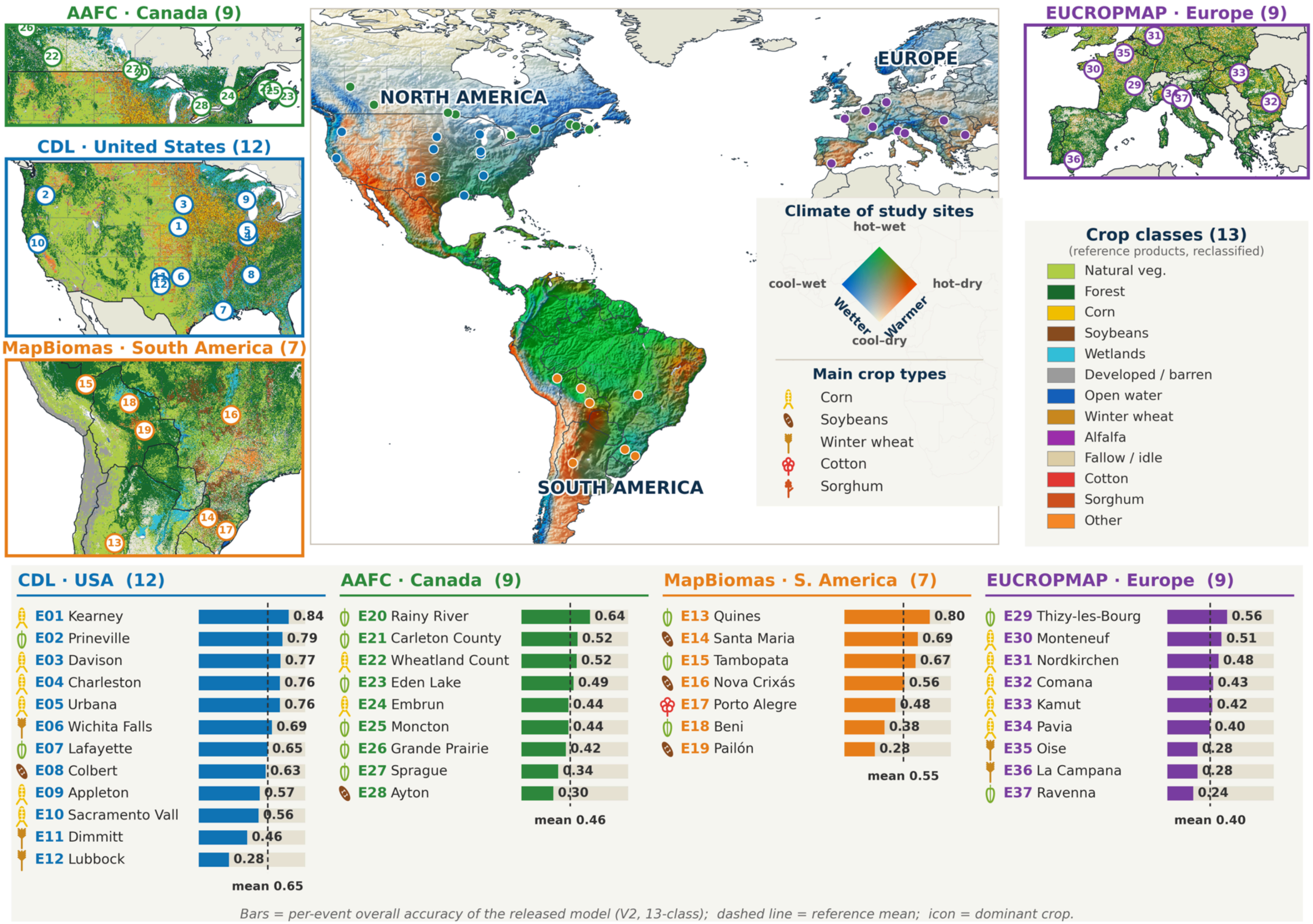

**Fig. 1. Study area and evaluation design.** The 37 out-of-distribution evaluation events span three continents and four national or continental crop-reference products (CDL, AAFC, MapBiomas, and EUCROPMAP). The central map shows event locations over a trivariate climatic base (mean temperature × precipitation, shaded by elevation); regional insets show each reference crop layer reclassified to the shared 13-class scheme, with events numbered E01–E37. The lower panel reports per-event overall accuracy of the released Prithvi-EO-2.0 model (V2, 13-class), grouped by reference product and ordered by descending accuracy. Country boundaries are from Natural Earth.

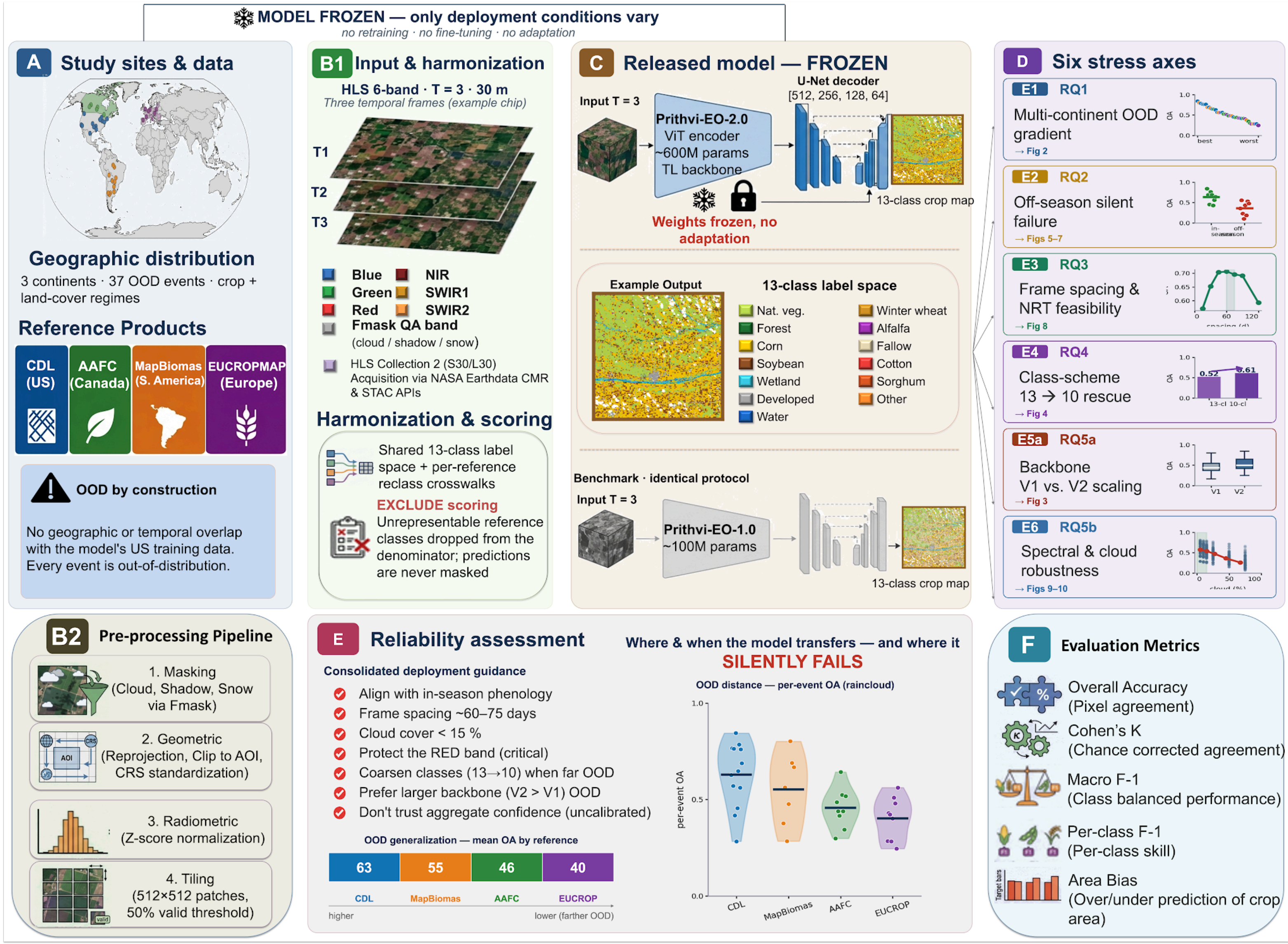

**Fig. 2. Evaluation workflow**. The released Prithvi-EO-2.0-600M-TL crop model is applied with frozen weights so that only the deployment conditions vary. Panels show (A) the 37 out-of-distribution events across three continents and four reference products, (B) the three-frame HLS input, its harmonization to a shared 13-class label space, and the preprocessing and exclude-from-denominator scoring, (C) the frozen model and its Prithvi-EO-1.0-100M benchmark, (D) the six controlled experiments corresponding to the research questions, (E) the consolidated reliability assessment and deployment guidance, and (F) the accuracy metric suite.

**Table 1. Event roster (37 events).** Event roster (37 events). Location, reference product, region, country, acquisition dates for the three HLS frames (T1–T3), bounding box (decimal degrees; W, S, E, N), area, and scene cloud and shadow fractions for each event. Events are grouped by reference product (ordered by descending mean overall accuracy) and, within each product, by descending V2 overall accuracy.

| ID | Event | Reference | Region | Country | T1 | T2 | T3 | Bounding box (W, S, E, N) | Area (km²) | Cloud (%) | Shadow (%) |
|---|---|---|---|---|---|---|---|---|---|---|---|
| 1 | Kearney, Nebraska | CDL | North America | USA | 2025-02-26 | 2025-05-31 | 2025-08-21 | -99.653, 40.407, -98.471, 41.287 | 9370.9 | 3.74 | 1.9 |
| 2 | Prineville, Oregon | CDL | North America | USA | 2024-04-23 | 2024-08-06 | 2024-11-07 | -121.077, 44.210, -120.265, 45.529 | 9006.7 | 4.78 | 0.39 |
| 3 | Davison, South Dakota | CDL | North America | USA | 2025-03-27 | 2025-06-30 | 2025-09-28 | -98.762, 43.353, -97.839, 43.993 | 5048.3 | 0.39 | 0.29 |
| 4 | Charleston, Illinois | CDL | North America | USA | 2024-03-19 | 2024-07-01 | 2024-10-05 | -88.154, 39.440, -87.600, 39.887 | 2242.6 | 1.89 | 1.67 |
| 5 | Urbana, Illinois | CDL | North America | USA | 2024-03-30 | 2024-07-13 | 2024-10-06 | -88.277, 40.175, -87.673, 40.502 | 1800.7 | 1.88 | 0.93 |
| 6 | Wichita Falls, Texas | CDL | North America | USA | 2024-02-08 | 2024-05-08 | 2024-08-16 | -98.969, 33.885, -98.428, 35.051 | 6293.7 | 0.11 | 0.09 |
| 7 | Lafayette, Louisiana | CDL | North America | USA | 2024-03-10 | 2024-06-13 | 2024-09-26 | -92.586, 29.843, -91.084, 30.544 | 11169.4 | 10.36 | 4.94 |
| 8 | Colbert, Alabama | CDL | North America | USA | 2024-03-28 | 2024-06-24 | 2024-09-20 | -87.908, 34.516, -86.692, 34.961 | 5410.5 | 1.02 | 0.79 |
| 9 | Appleton, Wisconsin | CDL | North America | USA | 2024-04-24 | 2024-07-26 | 2024-10-24 | -88.623, 43.723, -87.695, 44.811 | 8374.5 | 1.45 | 1.35 |
| 10 | Sacramento Valley, California | CDL | North America | USA | 2023-04-24 | 2023-07-25 | 2023-10-26 | -122.410, 37.875, -121.410, 39.614 | 16312.7 | 0.48 | 0.27 |
| 11 | Dimmitt, Texas | CDL | North America | USA | 2024-03-10 | 2024-06-13 | 2024-09-16 | -102.713, 34.033, -101.453, 34.845 | 10110.1 | 0.06 | 0.05 |
| 12 | Lubbock, Texas | CDL | North America | USA | 2024-03-11 | 2024-06-13 | 2024-09-11 | -102.360, 32.392, -101.723, 34.538 | 13520.4 | 0.54 | 0.43 |
| 13 | Quines, San Luis | MapBiomas | South America | Argentina | 2023-01-05 | 2023-03-18 | 2023-06-14 | -66.587, -32.187, -65.882, -30.731 | 10771 | 0.28 | 0.21 |
| 14 | Santa Maria, Rio Grande do Sul | MapBiomas | South America | Brazil | 2023-12-01 | 2024-01-31 | 2024-03-11 | -54.740, -29.223, -53.307, -27.513 | 26761.9 | 0.21 | 0.04 |
| 15 | Tambopata, Madre de Dios | MapBiomas | South America | Peru | 2023-05-15 | 2023-08-07 | 2023-10-11 | -70.351, -12.278, -69.617, -11.744 | 4736.1 | 0.21 | 0.2 |
| 16 | Nova Crixás, Goiás | MapBiomas | South America | Brazil | 2023-10-19 | 2023-12-16 | 2024-04-24 | -51.376, -16.245, -50.521, -15.291 | 9694.4 | 1.56 | 0.22 |
| 17 | Porto Alegre, Rio Grande do Sul | MapBiomas | South America | Brazil | 2024-02-02 | 2024-05-06 | 2024-08-14 | -51.970, -30.076, -51.203, -29.726 | 2873.5 | 0.05 | 0.04 |
| 18 | Beni, Bolivia | MapBiomas | South America | Bolivia | 2022-03-16 | 2022-06-14 | 2022-09-17 | -64.868, -14.862, -63.730, -13.723 | 15462.9 | 1.63 | 0.41 |

| ID | Event | Reference | Region | Country | T1 | T2 | T3 | Bounding box (W, S, E, N) | Area (km²) | Cloud (%) | Shadow (%) |
|---|---|---|---|---|---|---|---|---|---|---|---|
| 19 | Pailón, Santa Cruz | MapBiomas | South America | Bolivia | 2021-09-24 | 2021-12-28 | 2022-03-28 | -62.753, -18.122, -61.871, -17.192 | 9649.3 | 0.1 | 0.07 |
| 20 | Rainy River, Ontario | AAFC | North America | Canada | 2024-04-14 | 2024-08-01 | 2024-11-07 | -94.206, 48.544, -93.478, 49.158 | 3298.9 | 2.61 | 2.25 |
| 21 | Carleton County, New Brunswick | AAFC | North America | Canada | 2024-04-30 | 2024-07-21 | 2024-10-22 | -67.663, 46.216, -66.016, 46.976 | 10117.7 | 3.5 | 0.19 |
| 22 | Wheatland County, Alberta | AAFC | North America | Canada | 2022-03-22 | 2022-07-20 | 2022-10-18 | -113.723, 50.541, -112.437, 51.684 | 9996.6 | 0.69 | 0.42 |
| 23 | Eden Lake, Nova Scotia | AAFC | North America | Canada | 2024-04-27 | 2024-07-28 | 2024-10-20 | -63.165, 45.186, -61.548, 45.588 | 5392.3 | 0.12 | 0.11 |
| 24 | Embrun, Ontario | AAFC | North America | Canada | 2023-05-23 | 2023-08-23 | 2023-11-16 | -75.665, 45.045, -74.628, 45.751 | 6008.8 | 0.97 | 0.64 |
| 25 | Moncton, New Brunswick | AAFC | North America | Canada | 2024-04-10 | 2024-07-14 | 2024-10-18 | -65.917, 45.878, -64.985, 46.556 | 4962.8 | 0.17 | 0.17 |
| 26 | Grande Prairie, Alberta | AAFC | North America | Canada | 2023-05-02 | 2023-07-29 | 2023-10-12 | -119.444, 54.767, -117.880, 55.759 | 9281.1 | 0.29 | 0.27 |
| 27 | Sprague, Manitoba | AAFC | North America | Canada | 2024-04-21 | 2024-07-30 | 2024-10-30 | -96.038, 49.014, -95.367, 49.518 | 2558.8 | 0.72 | 0.63 |
| 28 | Ayton, Ontario | AAFC | North America | Canada | 2023-04-13 | 2023-07-10 | 2023-10-03 | -81.204, 43.792, -80.444, 44.367 | 3663.4 | 0.3 | 0.3 |
| 29 | Thizy-les-Bourgs, Rhône | EUCROPMAP | Europe | France | 2022-04-17 | 2022-07-16 | 2022-10-18 | 3.216, 45.867, 5.512, 46.259 | 7319.6 | 0.21 | 0.11 |
| 30 | Monteneuf, Morbihan | EUCROPMAP | Europe | France | 2022-05-21 | 2022-08-09 | 2022-11-12 | -2.648, 47.525, -1.721, 48.270 | 5299.4 | 0.2 | 0.21 |
| 31 | Nordkirchen, NRW | EUCROPMAP | Europe | Germany | 2022-03-18 | 2022-06-16 | 2022-10-09 | 6.228, 51.474, 8.864, 51.942 | 8381.2 | 0.77 | 0.69 |
| 32 | Comana, Giurgiu | EUCROPMAP | Europe | Romania | 2022-04-05 | 2022-07-14 | 2022-10-17 | 24.723, 43.502, 27.549, 44.793 | 30539 | 0.25 | 0.18 |
| 33 | Kamut, Békés | EUCROPMAP | Europe | Hungary | 2022-05-16 | 2022-07-23 | 2022-10-20 | 20.779, 46.938, 21.379, 48.018 | 4978.9 | 2.52 | 0.29 |
| 34 | Pavia, Lombardy | EUCROPMAP | Europe | Italy | 2022-04-28 | 2022-08-04 | 2022-11-04 | 9.742, 44.857, 10.914, 45.320 | 4503.6 | 2.8 | 0.92 |
| 35 | Oise | EUCROPMAP | Europe | France | 2023-03-01 | 2023-06-04 | 2023-09-07 | 2.454, 49.627, 2.865, 49.828 | 591.9 | 0.54 | 0.35 |
| 36 | La Campana, Seville | EUCROPMAP | Europe | Spain | 2022-04-16 | 2022-06-25 | 2022-09-28 | -6.212, 37.289, -4.479, 37.847 | 9047.5 | 0.58 | 0.46 |
| 37 | Ravenna, Emilia-Romagna | EUCROPMAP | Europe | Italy | 2023-02-12 | 2023-05-23 | 2023-08-26 | 11.542, 44.400, 12.295, 44.676 | 1781.3 | 0.74 | 0.61 |

## 2.3 Evaluation-event roster

We assembled 37 evaluation events spanning 12 countries, three continents, and four national or continental crop-reference products (Fig. 1; Table 1). Events were selected to place the model OOD by construction. None overlaps the model's United States fine-tuning data in the combination of region, cropping system, and season that defines the training distribution. The roster deliberately spans a gradient of distance from that distribution, from in-country United States sites, through Canadian and South American agricultural regions, to European cropping systems with distinct calendars and class nomenclatures. Each event is a contiguous scene processed at full extent (rather than isolated patches), so that accuracy is assessed at the map level.

Events are grouped by reference product: Cropland Data Layer (CDL; United States; n=12; Boryan et al., 2011), MapBiomas (South America; n=7; Souza et al., 2020), Agriculture and Agri-Food Canada Annual Crop Inventory (AAFC; Canada; n=9; Fisette et al., 2013), and EUCROPMAP (Europe; n=9; d'Andrimont et al., 2021). Each event is identified by a stable code (E01–E37), a locality name, its reference product and region, the acquisition dates of its three HLS frames, its bounding box and area, and its scene-level cloud and shadow fractions (Table 1), with a per-event phenological-phase and coverage audit in Table S5. For each event, the central frame (T2) year serves as the reference year, and the prediction is validated only against the reference layer for the matching year.

## 2.4 HLS acquisition and harmonization

For each event and each of the three target frames (T=3), we acquired HLS Collection 2 surface reflectance (the HLSL30 and HLSS30 products) through the NASA Earthdata portal. Because the two products can provide complementary spatial coverage on a given date, we selected the product with the greatest valid coverage of the event bounding box, determined from pixel availability rather than scene-level cloud metadata. The three frames of six bands were assembled into time-major 224×224 chips (channel order = frame × band), matching the input layout expected by the released model, and stitched across the full spatial extent of each event for inference. Cloud- and shadow-affected pixels were masked from the final products using the HLS quality assessment band and are also excluded from accuracy computation.

## 2.5 Reference products and shared-label-space validation

All comparisons are made in a shared 13-class label space, because the four reference products adopt incompatible class definitions. For each reference, we defined a reclassification crosswalk that maps its native classes to the model's 13 classes. Native classes with no representable counterpart are assigned to an excluded code. Validation is pixel-wise on the matched-year reference, resampled to the prediction grid by nearest-neighbor assignment. Qualitative per-event validation panels for each reference product are provided in Figures S9–S12.

The scoring convention is exclude-from-denominator, i.e., reference pixels that fall in an excluded (non-representable) class are removed from the accuracy denominator, but model predictions are never masked or relabeled. This ensures that the model is neither credited nor penalized for reference categories it cannot express, while every prediction it makes over adjudicable ground is scored. Accuracy is computed over the valid set of pixels for which both the prediction and the reference are defined.

### 2.6 Accuracy and calibration metrics

Per-event performance is summarized with a multi-metric suite that separates aggregate agreement from class-balanced skill, because the two diverge systematically under class imbalance. We report overall accuracy (OA), Cohen's κ (chance-corrected agreement), macro-averaged F1 (the unweighted mean of per-class F1, which weights minority crops equally with dominant classes), per-class F1, a crop-area bias (the ratio of predicted to reference area over a fixed crop set), and quantity disagreement (the total-area component of disagreement in the sense of Pontius) (Pontius and Millones, 2011). Overall accuracy and macro-F1 are reported together throughout because a model can achieve high OA on a few dominant classes while failing on the minority crops that macro-F1 exposes.

In unlabeled regions, the only quality signal is the confidence the model reports for its own predictions. We therefore complement the accuracy suite with calibration and uncertainty metrics for the off-season experiments. Expected calibration error (ECE) measures the agreement between predicted confidence and empirical accuracy. It is computed over 15 equal-width bins of the model's top-class softmax probability as the population-weighted mean absolute difference between confidence and accuracy within each bin. We have also computed a top-label Brier score on the same predictions (Guo et al., 2017). Moreover, a single-model softmax cannot express epistemic uncertainty, which is the model's uncertainty about whether an input lies within its training regime. We therefore estimate predictive uncertainty using Monte Carlo dropout (Gal and Ghahramani, 2016). Twenty stochastic forward passes are drawn with dropout active, and we report the predictive entropy of the pass-averaged class distribution and its across-pass variance. We note that the released model exposes dropout only in its prediction head. These Monte Carlo estimates, therefore, reflect head-level uncertainty and constitute a lower bound on the model's total epistemic uncertainty.

Each event configuration was evaluated with a single deterministic inference run. Because the model weights are fixed, and standard inference is effectively deterministic, the principal variation of interest is across events and deployment conditions rather than across random seeds. Moreover, pixel-wise agreement between individual Monte Carlo dropout predictions and the saved deterministic prediction ranged from 0.982 to 0.996. We therefore treat the cross-event distribution, shown directly in the per-reference results, as the variation of interest. The remaining stochasticity arises from two sources: Monte Carlo dropout sampling and the selection among available HLS acquisitions during data assembly.

### 2.7 Experimental design

Six controlled experiments stress the deployed model along the axes of the research questions. Each holds the released model fixed and varies one deployment condition. Three experiments use subsets of the 37 events, set by data availability and compute cost rather than by outcome. The off-season comparison (RQ2) requires a matched pair, meaning a correct-season and an off-season three-frame window at the same location, with the same reference product and year. Eight events met this condition, and one further off-season site was discarded because no correct-season counterpart could be constructed. The frame-spacing grid (RQ3) requires sufficient cloud-free acquisitions within each event bounding box (≥90% clear pixels after masking clouds, shadows, and snow). We found 20 such events, yielding 818 configurations across the requested arm lengths of 15, 30, 45, 60, 75, 90, and 120 days. The band-corruption and contamination experiment (RQ5b) costs 61 inference runs per event, so it was limited to 14 events. Each subset spans all four reference products, and Table S1 records which events were entered into which experiment.

*(i) Multi-continent OOD gradient (RQ1):* The released model is applied to all 37 events and scored against each event's regional reference in the shared label space. Per-event OA, macro-F1, and κ are aggregated by reference product along the CDL → MapBiomas → AAFC → EUCROPMAP gradient of increasing distance from the training distribution.

*(ii) Temporal-window and off-season behavior (RQ2)***:** For eight matched pairs, one three-frame window was aligned to the local growing calendar, and the other was drawn outside it. We compare OA, per-class F1, mean confidence, ECE, Brier score, and Monte Carlo predictive entropy between the two windows to test whether a corresponding loss of confidence accompanies the accuracy loss under seasonal misalignment, or whether the model fails silently. A high-confidence error mask (pixels that are both wrong and predicted with confidence at or above 0.50) localizes silent failures spatially.

*(iii) Frame spacing and near-real-time feasibility (RQ3)***:** A three-frame (T1-T2-T3) deployment requires selecting a central observation (T2) date and a target spacing between successive frames, which we refer to as the arm length (T1-T2 & T2-T3). Across 20 events, we varied both parameters to construct 818 three-frame configurations with requested arm lengths of 15, 30, 45, 60, 75, 90, and 120 days. Each flanking frame was taken from the nearest clear acquisition within 10 days of the target offset on the corresponding side of the central frame. The two realized gaps differ in most configurations, with a median of four days and a maximum of 20 days. Both realized offsets are reported for every configuration in Table S6. We estimated the relationship between overall accuracy and requested arm length using only configurations that fall within the local growing season. Off-season configurations collapse due to phenological misalignment rather than insufficient frame spacing, and averaging across both conditions would conflate distinct sources of error. All 34 configurations at the 120-day arm fell outside the growing season, so the in-season curve spans 15-90 days. Near-real-time deployment imposes a further restriction on this parameter space because a map produced during the growing season cannot draw on imagery acquired after the map date, which forces the central date into the recent past and compresses the arm length. We quantified the accuracy cost of this compression.

*(iv) Class-scheme reorganization (RQ4):* The model was fine-tuned on a 13-class scheme, but an operational user may post-hoc collapse the predicted labels to a coarser legend if the model cannot reliably support all 13 distinctions. We tested whether such a collapse reduces error. We derived the consolidated scheme from the model's empirical confusion structure by pooling confusion matrices across events and merging the class pairs with the highest off-diagonal exchange. We retained only merges that produce operationally meaningful classes. The merge was applied post hoc by collapsing each event's saved confusion matrix, which is equivalent to relabeling both the prediction and reference rasters without altering the inference.

*(v) Backbone generation under OOD (RQ5a):* Prithvi-EO-1.0-100M and Prithvi-EO-2.0-600M-TL are run on identical chips and references for every event. Their per-event OA is compared with the reference product to assess whether the in-distribution advantage of the larger, newer backbone is retained and how it scales with OOD.

*(vi) Spectral and contamination robustness (RQ5b):* For 14 events, we corrupt each of the six input bands in turn, under two fill strategies. Mean-fill replaces the band with its mean, removing its spatial information while keeping the input within the channel's normal range. The resulting OA loss measures how much the model depends on the information that the band carries. Zero-fill replaces the band with an out-of-range value, so the additional loss beyond mean-fill measures the extra damage caused by an out-of-range input rather than by the missing information. Separately, we inject synthetic cloud and shadow at controlled coverage fractions into either a single frame or all three frames and measure the degradation of OA together with the recoverability afforded by the three-frame composite.

## 3. Results

### 3.1 Accuracy declines across the multi-continent out-of-distribution gradient (RQ1)

We compared the model predictions against each event's regional reference product in the shared 13-class label space (Fig. 3, Table S1). Mean OA declined across the four reference products as the distance from the model's United States fine-tuning distribution increased. It fell from 0.646 on the CDL events (United States, n=12), to 0.552 on the MapBiomas events (South America, n=7), 0.457 on the AAFC events (Canada, n=9), and 0.402 on the EUCROPMAP events (Europe, n=9). Performance within each product varied widely. Several United States sites (Kearney 0.844, Prineville 0.79, Davison 0.77, Urbana 0.76) exceeded 0.75, whereas the weakest European events (Ravenna 0.24, La Campana 0.28, Oise 0.28) fell below 0.30.

The class-balanced metrics did not follow the same monotonic ordering. Macro-F1 averaged 0.336 across the 12 CDL events, 0.188 across the 7 MapBiomas events, 0.162 across the 9 AAFC events, and 0.245 across the 9 EUCROPMAP events. Cohen's κ followed the same order, at 0.531, 0.222, 0.175, and 0.306, respectively. The Canadian AAFC events returned the lowest

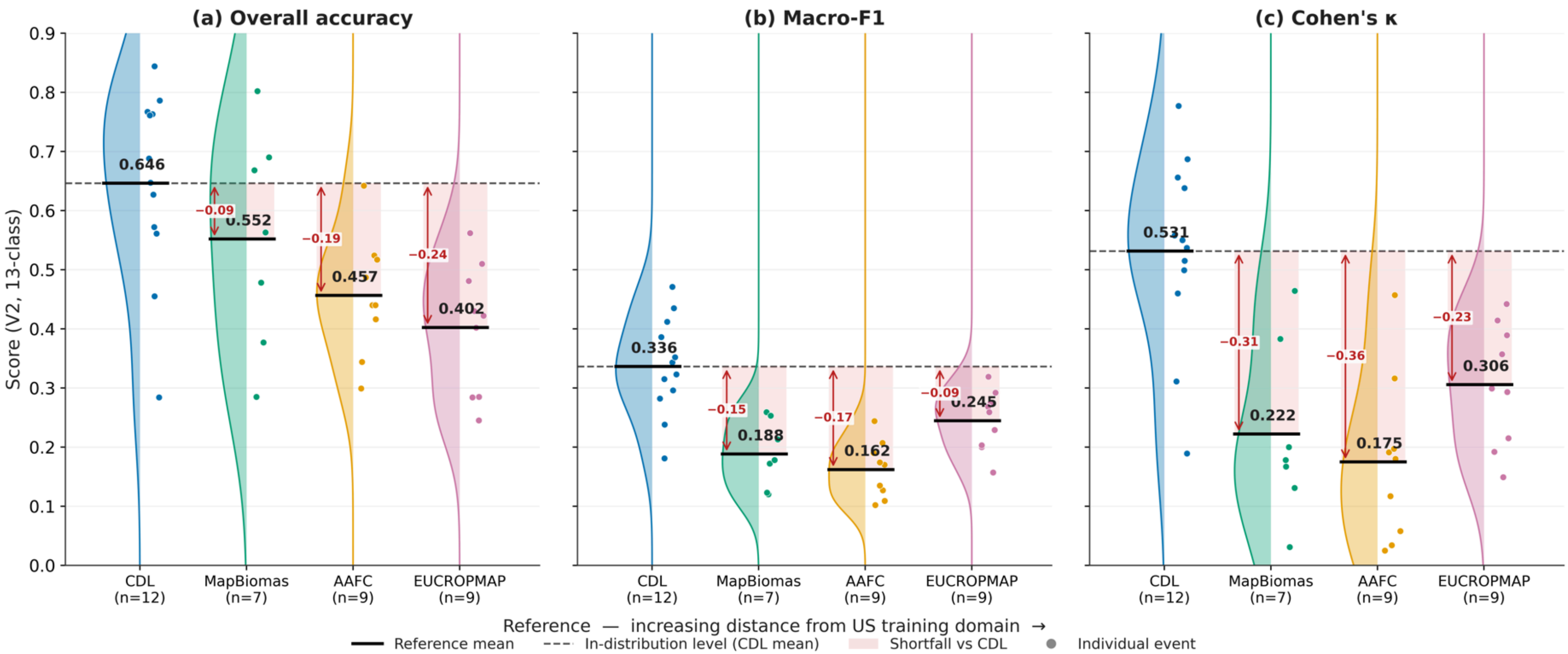

**Fig. 3. Out-of-distribution accuracy gradient across reference products.** Distribution of per-event (a) overall accuracy, (b) macro-F1, and (c) Cohen's κ for the released model (V2, 13-class) across the four reference products, ordered by increasing distance from the US training domain. Half-violins show the per-reference distribution, points are individual events, and the solid bar is the reference mean; the dashed line marks the in-distribution (CDL) mean, and shaded bands give each reference's shortfall relative to it. Overall accuracy declines monotonically (0.646 → 0.402), whereas macro-F1 and κ show the same overall drop but not a monotonic ordering, with AAFC lowest on the class-balanced metrics despite a mid-range overall accuracy.

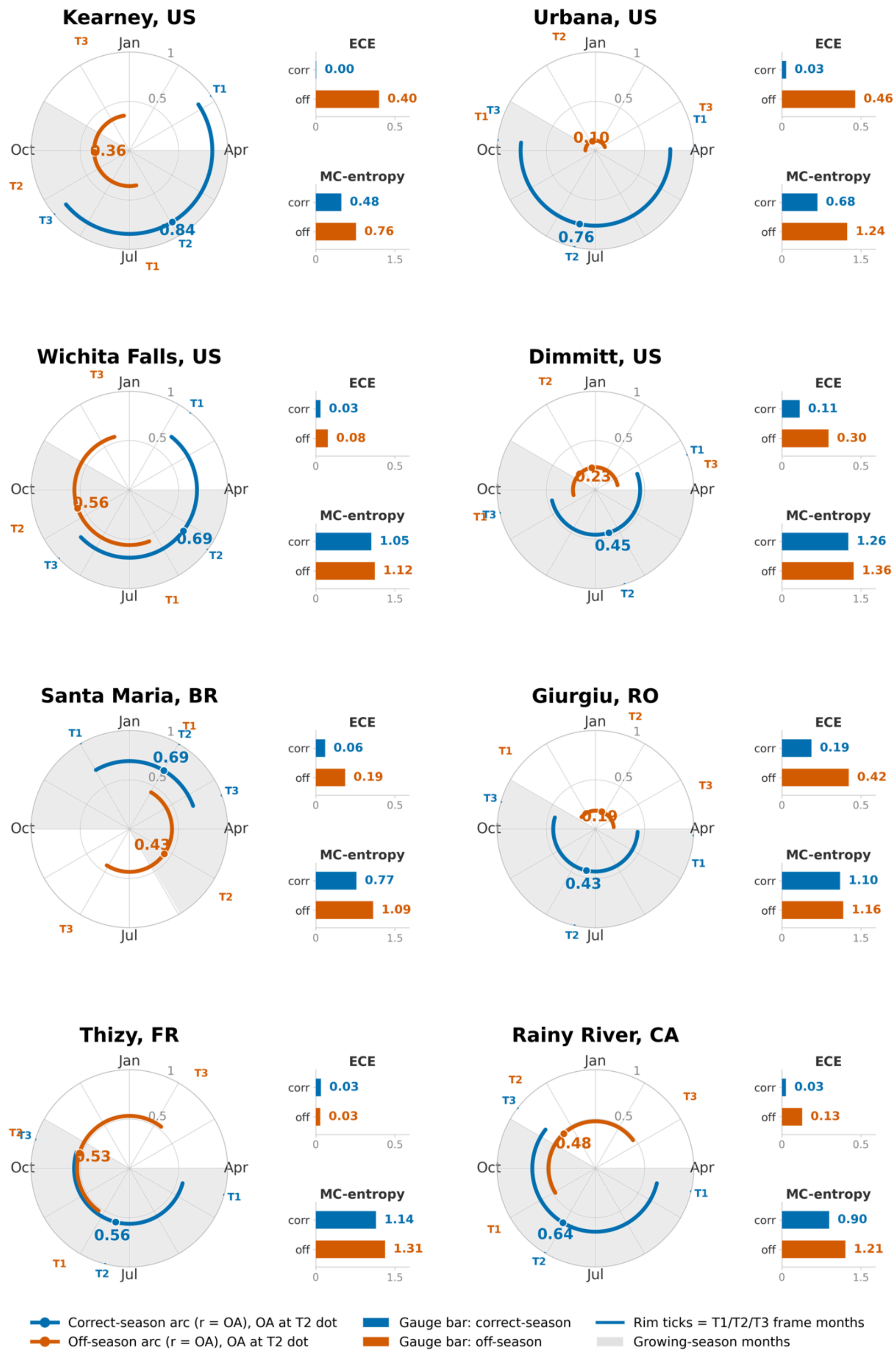

**Fig. 4. Phenology clock: accuracy and calibration under seasonal mistiming.** For each of the eight correct-versus-off-season pairs, a circular annual calendar shows the correct-season (blue) and off-season (orange) windows with the three HLS frames marked on the rim and the growing season shaded; arc radius encodes overall accuracy, and the paired gauges report expected calibration error (ECE) and MC-dropout predictive entropy for each window. Off-season windows collapse overall accuracy while mean confidence stays high, raising ECE in seven of eight pairs and predictive entropy in all eight. MC-dropout uses 20 stochastic passes with head dropout (p=0.1) and is a lower bound on epistemic uncertainty.

macro-F1 and κ despite a mid-range OA, while the EUCROPMAP events, lowest on OA, recovered to the second-highest macro-F1. This re-ordering reflects the class composition of the scenes rather than the quality of transfer. A small number of land-cover classes dominate AAFC scenes, so OA is driven by agreement on those dominant classes. At the same time, minority crops that macro-F1 weights equally are missed, lowering the balanced metrics. EUCROPMAP scenes contain a more even mix of classes, so agreement is distributed across classes, and the two metric families give closer values. The ranking of two deployments can invert depending on whether dominant-class agreement or per-class skill is measured. We therefore report overall accuracy and macro-F1 together throughout, and the same distinction recurs in the class-scheme and temporal analyses below.

Overall, the model generalized only partially beyond its training distribution. Aggregate accuracy decreased with distance from the United States training distribution, but the best out-of-distribution events remained well below the in-distribution ceiling. The decrease was steepest for minority crop classes rather than dominant land-cover classes, and the between-product ranking depended on the metric used. These results frame the two questions the remaining experiments address. Section 3.2 examines what drives the loss of accuracy beyond geographic distance, and Sections 3.3 and 3.4 examine which deployment choices recover accuracy.

### 3.2 Off-season deployment produces silent, confident failure (RQ2)

We separated the effect of the temporal observation window from that of geographic distance by constructing matched pairs of three-frame windows for eight sites. Both windows cover the same location, reference product, and year. One window was aligned with the local growing calendar (growing season), and the other was drawn outside it (off-season). Off-season deployment reduced OA at all eight pairs, and by a large margin at six of them. OA fell from 0.844 to 0.359 at Kearney, from 0.761 to 0.102 at Urbana, from 0.690 to 0.433 at Santa Maria, from 0.430 to 0.187 at Giurgiu, from 0.455 to 0.229 at Dimmitt, and from 0.642 to 0.477 at Rainy River. The two smaller reductions occurred at Wichita Falls (0.688 to 0.559) and Thizy (0.562 to 0.532). These were the pairs with off-season windows least displaced from the growing season. This pattern is consistent with displacement from local phenology driving the loss, rather than the calendar label of the window.

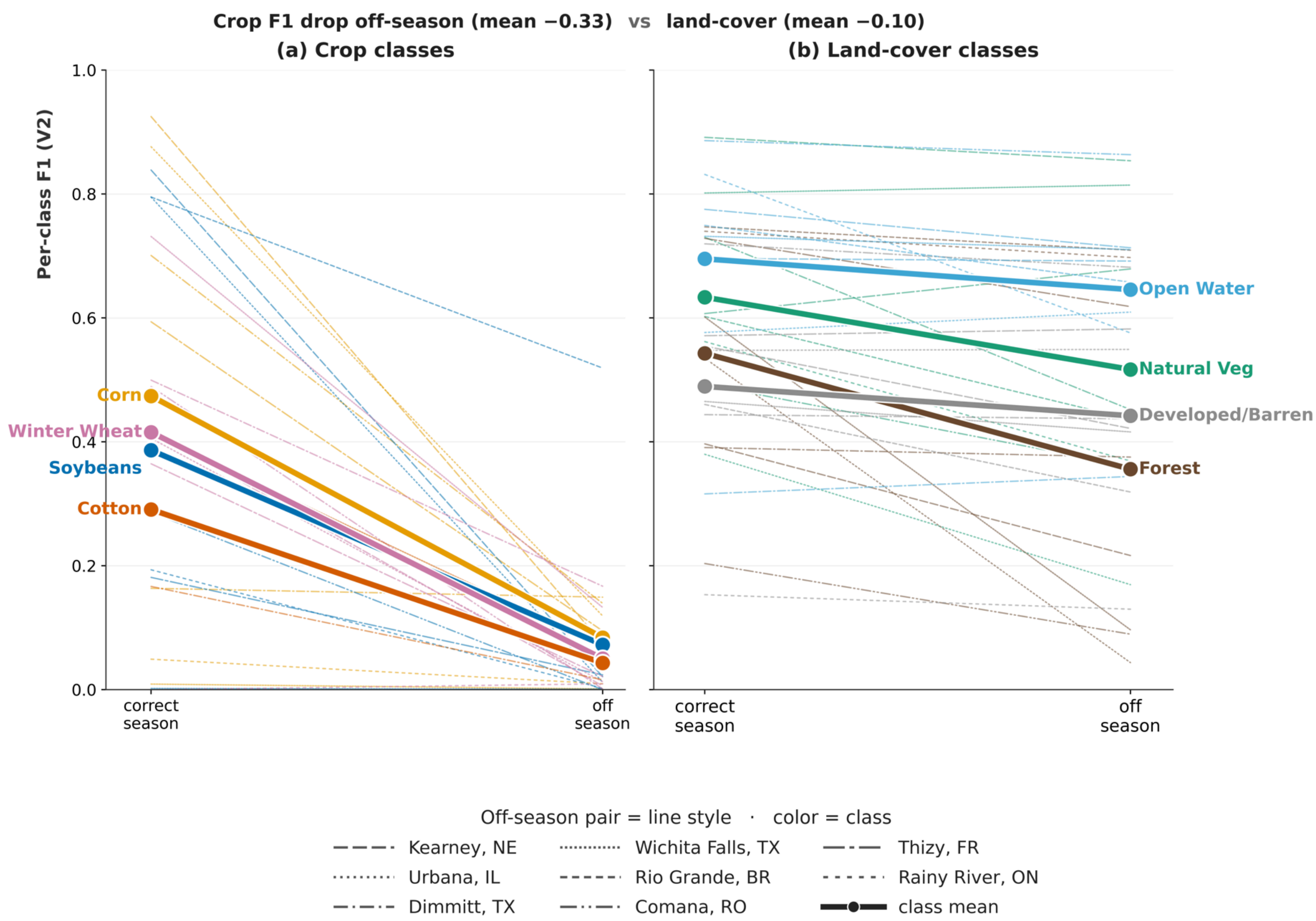

**Fig. 5. In-season vs off-season per-class F1.** Per-class F1 of the released model (V2) in the correct-season and off-season windows for the eight temporal pairs, separated into crop classes and land-cover classes; thin lines are individual events and bold lines are class means. Crop-class F1 collapses off-season, retaining only ~12–19% of its in-season value (mean change −0.33), whereas land-cover classes retain 66–93% (mean change −0.10).

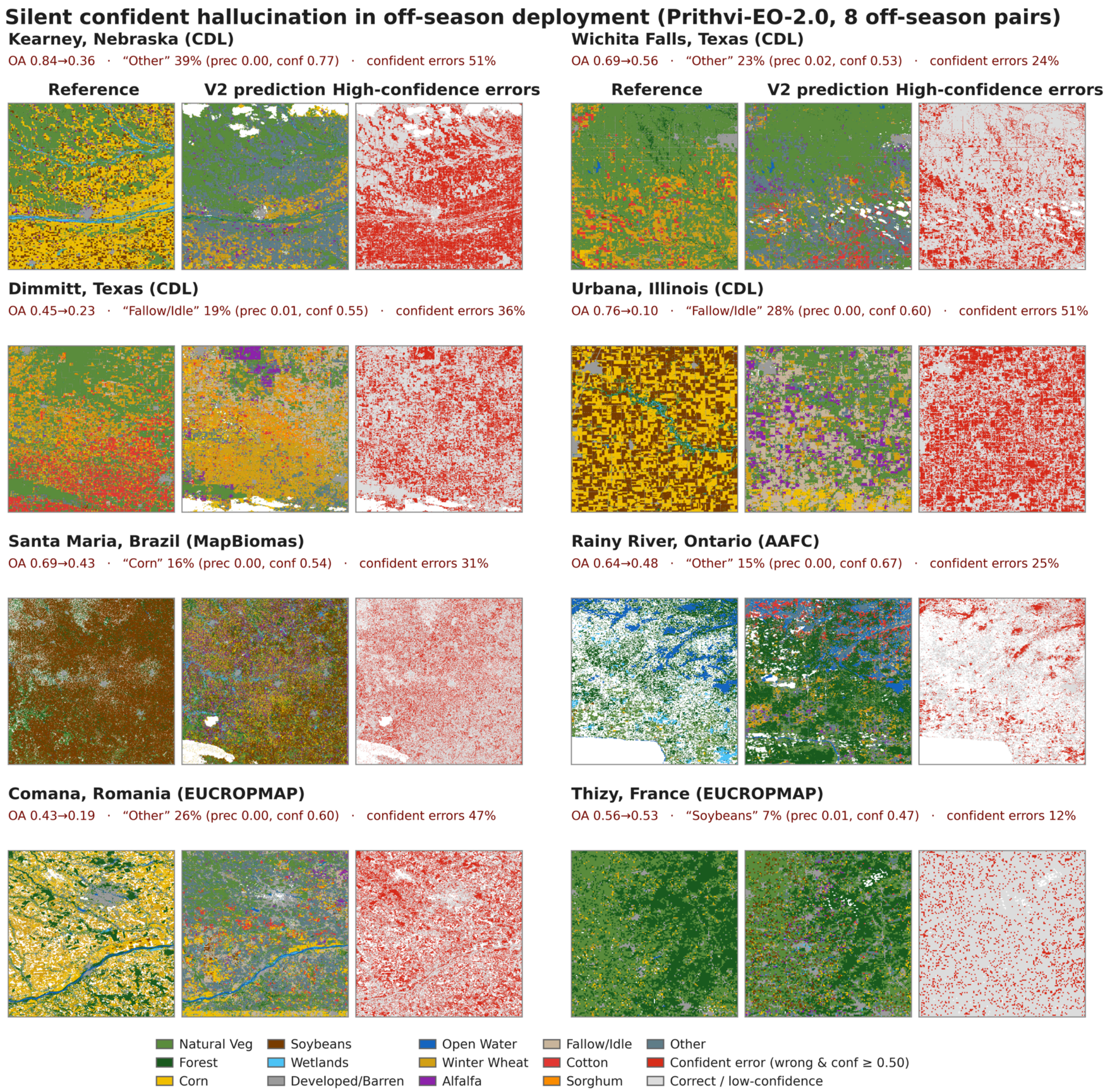

**Fig. 6. Silent confident hallucination in off-season deployment.** For all eight off-season pairs, the aligned maps show the reference labels, the V2 prediction, and the high-confidence error mask (pixels that are both wrong and predicted with confidence $\geq 0.50$). Annotations give the correct-season → off-season overall accuracy, the most-hallucinated class with its predicted-area fraction, precision, and mean confidence, and the scene share occupied by high-confidence errors. Off-season, the model assigns crop classes to large fractions of the scene at high confidence but near-zero precision, leaving much of the scene confidently wrong.

Mean predicted confidence remained high despite reduced accuracy in six of the eight pairs. At Urbana, confidence dropped only from 0.78 to 0.57, while OA fell by two-thirds. The model was therefore almost as confident when its predictions were mostly wrong as when they were mostly correct. ECE rose off-season at seven of the eight pairs, from 0.004 to 0.401 at Kearney, from 0.027 to 0.463 at Urbana, from 0.059 to 0.185 at Santa Maria, and from 0.187 to 0.423 at Giurgiu (Fig. 4; Table 2; Fig. S8). The only exception was Thizy, with ECE that was essentially unchanged (0.032 to 0.028) and which was also the mildest shift in the set. Monte-Carlo dropout predictive entropy, in contrast, rose off-season at all eight pairs, including Thizy, and by a large margin at the most affected pairs (0.484 to 0.763 at Kearney, 0.676 to 1.239 at Urbana). Deterministic confidence failed to flag the accuracy loss at seven of the eight pairs, whereas the Monte-Carlo entropy flagged it at all eight. We examine the reason for this in Section 4.2.

The per-class structure of the failure shows what the model does when the phenological signal is absent (Figs. 5 and 6). The model did not abstain or lower its confidence. It assigned incorrect crop labels to large fractions of the scene at high confidence. At Kearney, the "other" class expanded from under 1% of the scene in the correct season to 39% off-season, at a precision of 0.00 and a mean confidence of 0.77. That confidence exceeded the value assigned by the smaller V1 model to the same class (0.46). At Urbana, the fallow or idle class covered 28% of the scene off-season at zero precision and a mean confidence of 0.60, while natural vegetation and winter wheat expanded to 34% and 11%. Across the eight pairs, high-confidence errors, defined as pixels that are both wrong and predicted with confidence at or above 0.50, occupied between 12% (Thizy) and 51% (Kearney and Urbana) of the scene. The per-class results show that this behavior is specific to crops (Fig. 5). In the off-season, crop-class F1 fell to between 12% and 19% of its in-season value (mean change: -0.33). Land-cover classes such as open water, natural vegetation, and developed or barren land retained 66% to 93% of their in-season F1 (mean change -0.10). Crops are distinguished from one another by their phenological trajectories over the season. Hence, a window drawn outside the growing season removes the signal on which crop discrimination depends, while land-cover classes that are separable on a single date are largely unaffected.

Overall, the temporal window is a first-order control on the model's accuracy, with an effect comparable in magnitude to geographic distance. Seasonal misalignment produces a failure that the model's confidence does not flag. Part of the cross-continent accuracy gradient in Section 3.1 may therefore be a phenological-window effect rather than a pure geographic-transfer effect. An operational user cannot rely on the model confidence to detect a mistimed deployment, which motivates the reliability-aware guidance in Section 4.4.

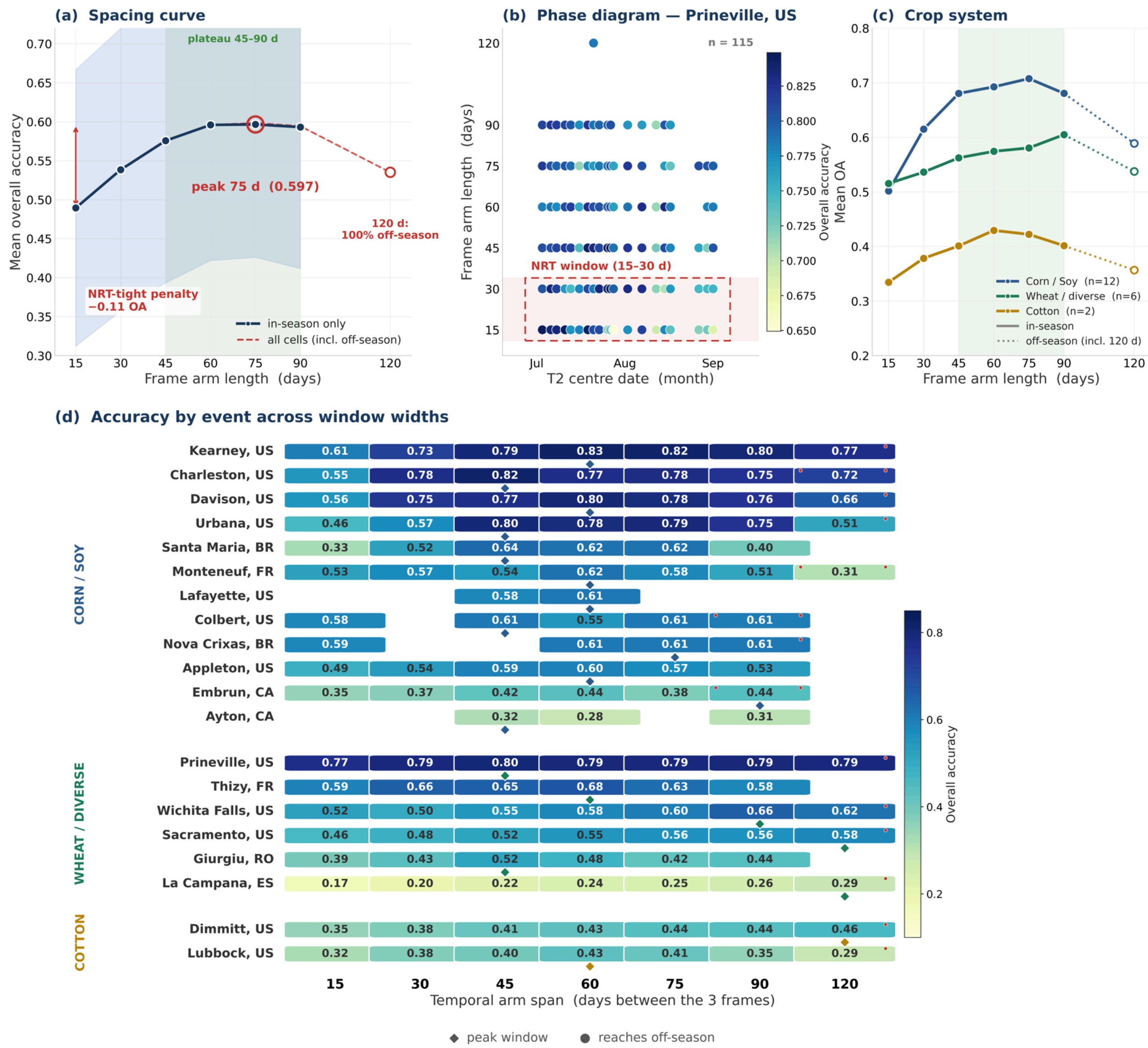

**Fig. 7. Frame spacing and near-real-time feasibility.** (a) Mean overall accuracy versus frame arm length (818-cell gap-grid, 20 events), over in-season cells only (solid) and all cells (dashed); accuracy plateaus across 45–90 days, peaks near 75 days, and near-real-time-tight arms (<30 days) incur a ~0.11 penalty. (b) Phase diagram for a representative event (Prineville): overall accuracy as a function of T2 center date and arm length, with the near-real-time window boxed. (c) Mean overall accuracy versus arm length by crop system. (d) Overall accuracy by event across window widths, grouped by crop system; the peak window per event is marked.

**Table 2. Calibration and silent-failure metrics (eight off-season pairs, V2).** Overall accuracy, mean confidence, expected calibration error, top-label Brier score, and MC-dropout predictive entropy for the correct-season and off-season windows of each pair. Off-season windows collapse overall accuracy while mean confidence remains high, so calibration error rises (seven of eight pairs) and predictive entropy increases (all eight). MC-dropout entropy uses 20 stochastic passes with head dropout (p=0.1) and is a lower bound on epistemic uncertainty.

| Event | Reference | Window | OA | Mean conf. | ECE | Brier | MC-entropy |
|---|---|---|---|---|---|---|---|
| Kearney, Nebraska | CDL | Correct-season | 0.844 | 0.846 | 0.004 | 0.092 | 0.484 |
| | | Off-season | 0.359 | 0.76 | 0.401 | 0.373 | 0.763 |
| Urbana, Illinois | CDL | Correct-season | 0.761 | 0.779 | 0.027 | 0.128 | 0.676 |
| | | Off-season | 0.102 | 0.565 | 0.463 | 0.329 | 1.239 |
| Wichita Falls, Texas | CDL | Correct-season | 0.688 | 0.658 | 0.03 | 0.176 | 1.052 |
| | | Off-season | 0.559 | 0.635 | 0.076 | 0.201 | 1.119 |
| Dimmitt, Texas | CDL | Correct-season | 0.455 | 0.568 | 0.113 | 0.222 | 1.258 |
| | | Off-season | 0.229 | 0.524 | 0.295 | 0.293 | 1.361 |
| Santa Maria, RS | MapBiomas | Correct-season | 0.69 | 0.749 | 0.059 | 0.168 | 0.773 |
| | | Off-season | 0.433 | 0.619 | 0.185 | 0.266 | 1.086 |
| Comana, Giurgiu | EUCROPMAP | Correct-season | 0.43 | 0.616 | 0.187 | 0.26 | 1.104 |
| | | Off-season | 0.187 | 0.61 | 0.423 | 0.33 | 1.165 |
| Thizy-les-Bourgs | EUCROPMAP | Correct-season | 0.562 | 0.589 | 0.032 | 0.22 | 1.142 |
| | | Off-season | 0.532 | 0.56 | 0.028 | 0.218 | 1.315 |
| Rainy River, ON | AAFC | Correct-season | 0.642 | 0.666 | 0.025 | 0.198 | 0.898 |
| | | Off-season | 0.477 | 0.606 | 0.129 | 0.259 | 1.205 |

### 3.3 Frame spacing sets a near-real-time accuracy ceiling (RQ3)

We examined how frame spacing affects accuracy, the binding constraint for near-real-time deployment. Across 20 events, we constructed 818 in-season three-frame configurations with requested arm lengths of 15, 30, 45, 60, 75, and 90 days (all 120-day configurations fell off-season; Fig. 7; Fig. S4).

Mean OA increased with arm length, reached a plateau, and then decreased. It rose from 0.489 at a 15-day arm to a peak of 0.597 at a 75-day arm, remained within a narrow band across the 45-to-90-day range, and fell below the plateau at arms shorter than 30 days. Arms tighter than 30 days incurred a penalty of about 0.11 in overall accuracy relative to the plateau. The model was fine-tuned on a widely spaced window that samples distinct phenological stages, so a tightly spaced window drawn from a single recent interval presents three near-duplicate observations and withholds the phenological contrast the model was trained to use.

The optimal arm length varied with the cropping system (Figs. 7c and 7d). Corn and soybean events peaked at a 75-day arm, achieving the highest accuracy of any cropping system (0.708) and tolerating tighter windows relatively well. The wheat and mixed-cropping events peaked at a 90-day arm (0.605), indicating that longer windows are beneficial for systems with more extended phenological development. The two cotton and sorghum events peaked earlier at the 60-day arm (0.429) and were the most sensitive to spacing. The per-event results show that the peak arm length is a stable property of each event and crop system, rather than an averaging artifact, as indicated by the peak window marked for each event in the heatmap (Fig. 7d).

Overall, in-season near-real-time deployment of the model is feasible but bounded. Accuracy is preserved across the 45-to-90-day plateau, so a user with moderate latency tolerance loses little. A window compressed to an arm's length below 30 days sits about 0.11 below the accuracy plateau, and both the penalty size and the optimal location depend on the cropping system. Together with the seasonal-alignment result from Section 3.2, these results showed that accurate deployment depended on phasing the window to the growing season and spacing its frames roughly two to three months apart within that season.

### 3.4 A data-derived class merge improves accuracy at every event (RQ4)

The class-composition effect in Section 3.1 indicated that the native 13-class scheme may be finer than the model can support under out-of-distribution conditions. We therefore tested consolidation into a 10-class operational scheme by merging the pairs that the model most often confused. These merges were forest with wetlands, alfalfa with fallow or idle cropland, and sorghum with the "other" class (Fig. 8; Fig. S2; Table S3). The merge was defined empirically and applied post hoc by collapsing each event's saved confusion matrix, equivalent to relabeling the rasters without altering inference.

The reorganization raised OA at all 37 events by an average of 8.4 percentage points. Because merging transfers off-diagonal counts onto the diagonal, OA cannot decrease, and therefore, only the magnitude is informative. That magnitude increased with distance from the training distribution, with regional-mean improvements of 0.020 across CDL events, 0.040 across EUCROPMAP, 0.111 across MapBiomas, and 0.192 across Canadian AAFC. The AAFC events gained the most because they had the lowest class-balanced scores (Section 3.1), and the consolidation removes the minority-class confusions that had lowered those scores. The largest individual gains came from events dominated by the merged classes. Moncton improved by 0.42, and four further Canadian and South American events improved by more than 0.25.

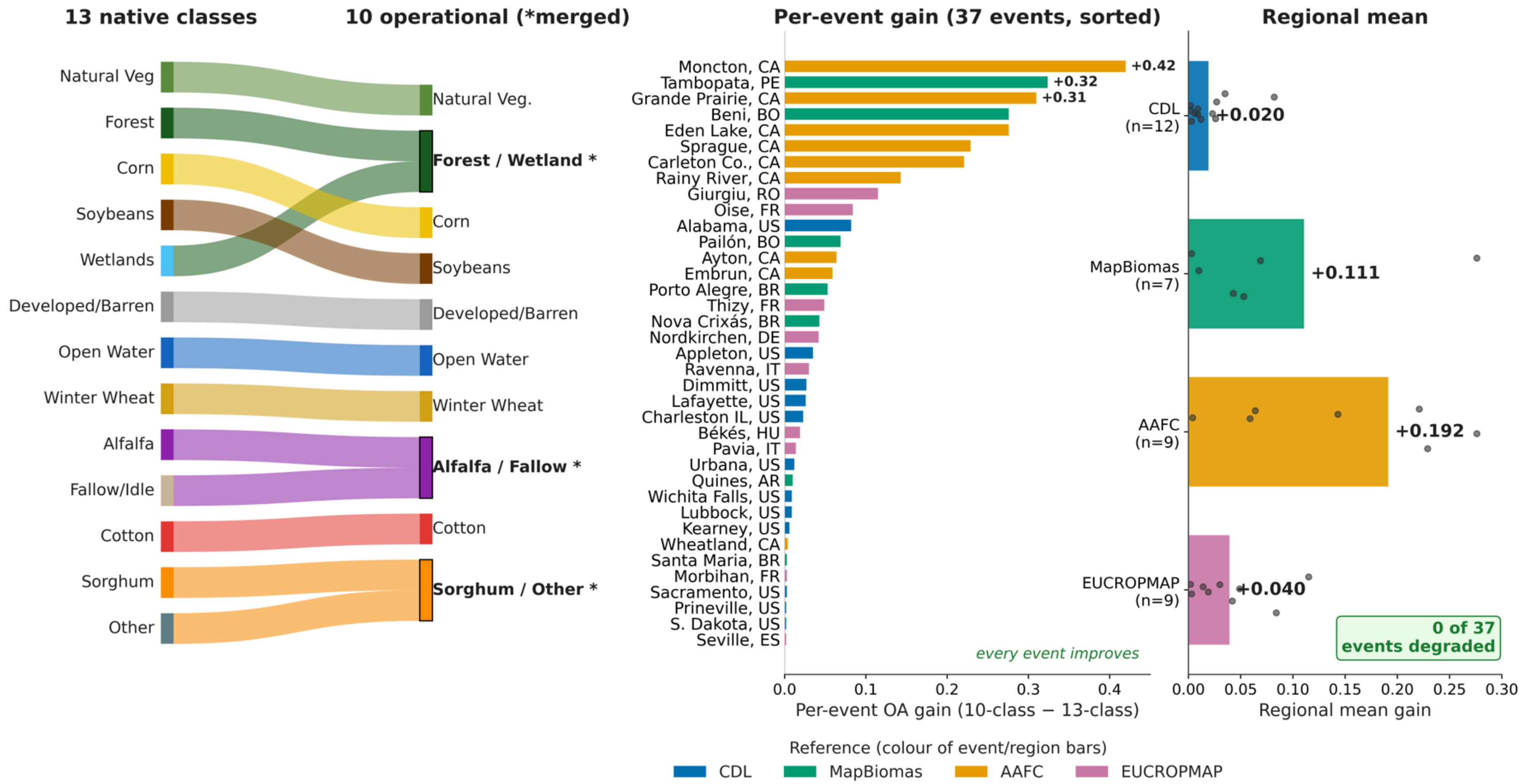

**Fig. 8. Class-scheme reorganization (13 -> 10).** Effect of consolidating the native 13-class legend into a 10-class operational scheme (merging Forest with Wetlands, Alfalfa with Fallow/Idle, and Sorghum with Other). The alluvial diagram shows consolidation; the center panel shows the per-event change in overall accuracy (10-class − 13-class), sorted and colored by reference; and the right panel shows the regional mean change. Coarsening the legend improves overall accuracy at every event (0 of 37 degraded) and recovers the greatest accuracy where the model is furthest from the distribution (regional mean ΔOA: CDL +0.020, EUCROPMAP +0.040, MapBiomas +0.111, AAFC +0.192).

Class-balanced metrics test whether the merge recovers structure or merely relabels errors. Macro-F1 rose from 0.244 to 0.313, a mean gain of 0.070, and improved at all 37 events. Cohen's κ rose from 0.331 to 0.394, a mean gain of 0.063, and improved at 35 of them. The two exceptions (La Campana and Sacramento) fell by 0.002 and 0.001, where OA increased by only 0.002, a consequence of κ discounting expected agreement when classes are fewer and larger. A merge that resolves almost no confusion can therefore leave κ fractionally lower. The results showed that coarsening the legend to the granularity supported by the model is an operational gain rather than a resolution-for-accuracy trade-off.

Overall, the results indicated that a substantial portion of the model's apparent error was attributable to class distinctions it could not support outside its training distribution. This component was recoverable at no cost by matching the operational legend to the model's demonstrated resolution. This was the second user-controllable adjustment, alongside window alignment, that improved accuracy, and it did so most where the model was otherwise weakest.

### 3.5 The V1-to-V2 backbone upgrade helps more as the shift grows (RQ5a)

We compared the released Prithvi-EO-2.0-600M-TL with its predecessor, Prithvi-EO-1.0-100M, using identical chips and references to isolate the effect of the backbone upgrade (Fig. 9, Fig. S1, Table S2). The newer, larger backbone improved OA on average, with gains widening as the distance from the training distribution increased. The mean V2-minus-V1 gain in OA was 0.055 for the CDL events, 0.068 for MapBiomas, 0.041 for AAFC, and 0.105 for the European EUCROPMAP events. The class-balanced metrics followed the same order, with the largest gains in κ and mean IoU also observed at the European events (Fig. S1).

The V2 advantage was not confined to the in-distribution benchmarks. The near tie in the CDL events (0.055) widened to a clear margin in the European events (0.105), indicating that the larger pretrained backbone yields a more transferable representation. At arm lengths below 30 days, V1's mean OA fell to 0.11–0.23, and many windows at these arms failed almost completely. In contrast, V2 retained usable accuracy across the same windows, which is the largest V1-to-V2 difference observed in the study. Overall, we found that the V1-to-V2 upgrade reduces but does not eliminate the OOD penalty. Even with the larger backbone, accuracy under strong geographic and seasonal shifts remained far below the in-distribution ceiling reported in Section 3.1.

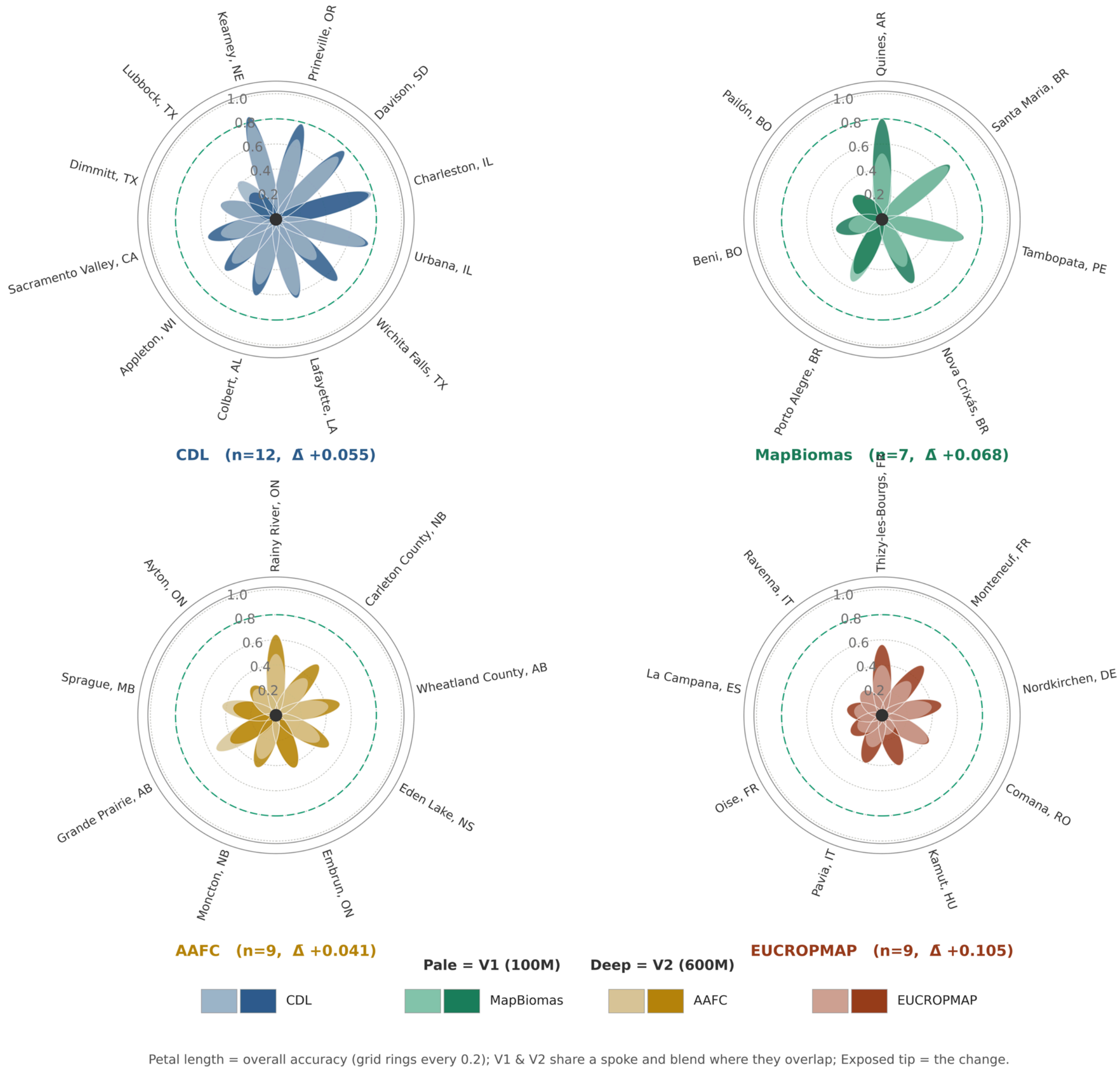

**Fig. 9. Backbone comparison across model generations (Prithvi-EO-1.0 → 2.0).** Per-event overall accuracy for the 100M-parameter V1 (pale petals) and the released 600M-parameter V2 (deep petals), as one radial rose per reference product; petal length encodes overall accuracy, and the exposed deep-petal tip is the V2 gain. Panel labels give the mean V2−V1 difference per reference (ΔOA = +0.055, +0.068, +0.041, and +0.105). The V2 advantage widens with domain shift, reaching its maximum at the European EUCROPMAP sites.

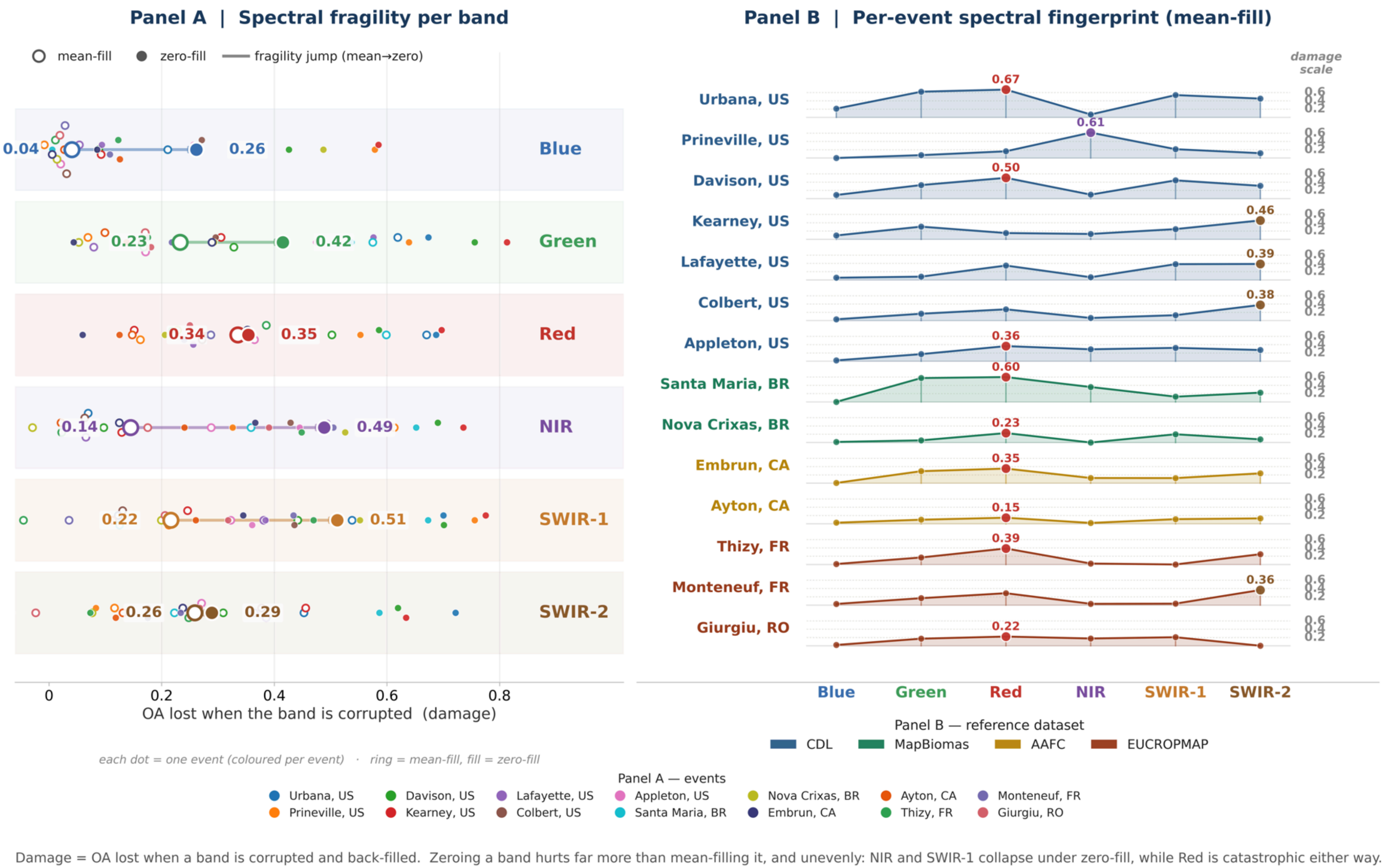

**Fig. 10. Spectral fragility under band corruption.** (a) Per-band loss of overall accuracy (V2, 14 events) under mean-fill imputation (open rings) and zero-fill (filled dots), with the connecting line showing the fragility jump between them. (b) Per-event spectral fingerprint under mean-fill across the six HLS bands, colored by reference product. Under mean-filling, the Red band is the most damaging to lose in every region; zeroing a band is far more harmful than mean-filling it, with the collapse concentrated in NIR and SWIR-1.

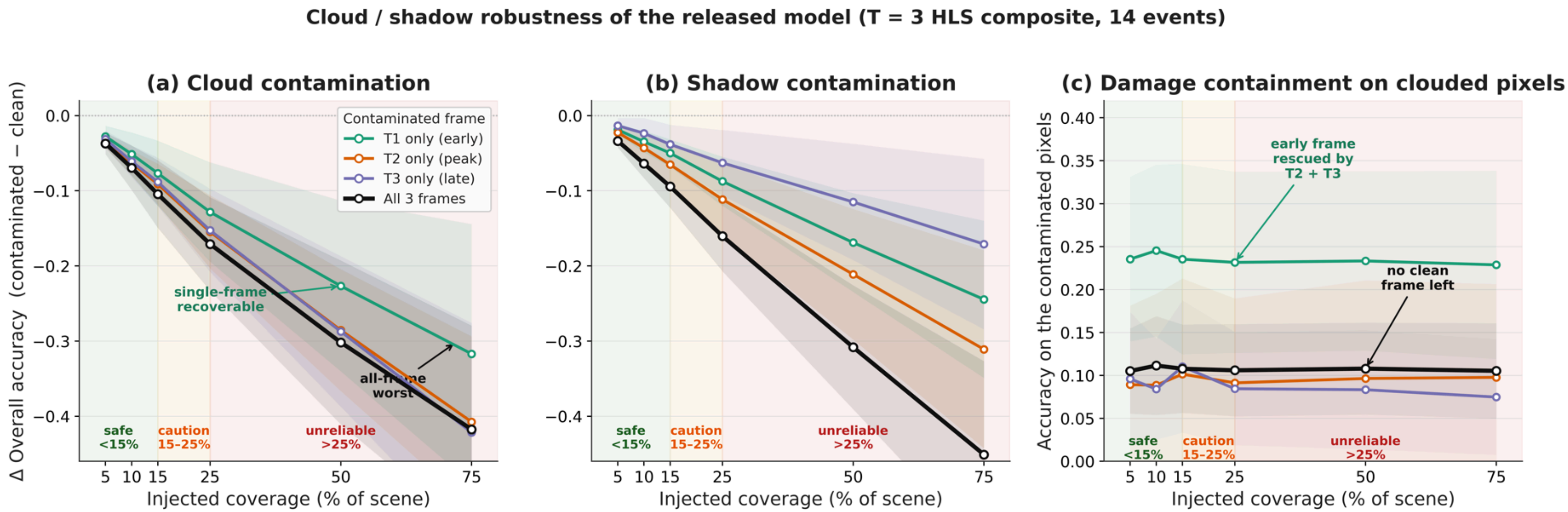

**Fig. 11. Cloud and shadow robust envelope.** Change in overall accuracy (V2, contaminated − clean, 14 events) versus injected contamination coverage, for (a) cloud and (b) shadow applied to a single frame or all three; lines are means and bands the inter-event spread, with safe (<15%), caution (15–25%), and unreliable (>25%) zones marked. (c) Accuracy on the contaminated pixels themselves. Accuracy degrades monotonically with coverage; all-frame contamination is the most damaging, while single-frame contamination is partially recovered in the T=3 composite.

### 3.6 Band dependence and contamination robustness (RQ5b)

We characterized the model's sensitivity to band corruption and cloud/shadow contamination across 14 events (refer to Section 2.7 for more details). Under mean-fill, the red band was the most damaging to lose across all regions, reflecting its central role in vegetation spectral contrast, which is the primary basis for distinguishing crops from one another and from non-crop land cover. Zero-filling a band was more harmful than mean-filling it in every case, with additional damage concentrated in the near-infrared and shortwave-infrared-1 bands (Fig. 10a). These results indicate that the model can partly compensate for a plausibly imputed band but not for an out-of-range one. The near-infrared and shortwave channels must carry physically sensible values for the model to remain stable, even though the red band carries the most discriminative crop information. The per-event fingerprints showed consistent ranking across events, with red-band sensitivity dominant in nearly every event's profile (Fig. 10b; Fig. S6; Table S7).

For the contamination experiment, synthetic clouds and shadows were injected at controlled coverage fractions into either a single frame or all three frames. All-frame contamination was more damaging than single-frame contamination at every coverage level (Fig. 11; Fig. S5; Table S8). The single-frame damage was largely recoverable because the two remaining clean frames preserved the temporal signal, but accuracy collapsed when all three frames were affected (Fig. 11c). Expressed as an operational contamination budget, the three-frame composite remained reliable below 15% scene coverage, entered caution between 15% and 25%, and became unreliable above 25%.

Together, the spectral and contamination experiments identify the red band and the multi-frame temporal composite as the model's two principal input dependencies. The red band contributed the most to crop discrimination, and the three-frame composite provided resilience to single-frame cloud or shadow contamination. These thresholds are consolidated with the temporal and class-scheme guidance in Section 4.4.

## 4. Discussion

### 4.1 The multi-continent accuracy gradient is substantially a phenological window effect

Overall accuracy declined from 0.646 at in-distribution CDL sites to 0.402 at European EUCROPMAP sites, which is well below the 80.3% OA reported for EUCROPMAP (d'Andrimont et al., 2021) and the >82% typically reached by supervised classifiers within their training region (Sun et al., 2020). This gap is comparable to cross-region losses in the pre-foundation-model literature (Ma et al., 2024), indicating that backbone pretraining alone does not eliminate the geographic generalization gap. This degradation is not specific to Prithvi-EO-2.0. Prior work reports that foundation model embeddings competitive in the United States collapse under leave-one-country-out transfer (Ma et al., 2026; Adjei, 2026). Foundation model OOD degradation is therefore a general property rather than a feature of one model (Chang et al., 2025; Sachdeva et al., 2024), and its magnitude scales with the geographic and climatic diversity of the evaluation set.

Much of this apparent geographic decay, however, reflects misalignment between the observation window and local phenology rather than pure spatial transfer. Our matched-pairs experiment isolates this by holding location fixed while varying only the window. A within-region natural experiment at two Brazilian sites, one correctly phased and one mis-phased, shows the same contrast (Fig. S7). This aligns with prior work that attributes cross-region drops to phenological discrepancies (Wang et al., 2022; Nyborg et al., 2022) and shows that temporal shifts can exceed regional shifts in severity (Fang et al., 2026; Ma et al., 2026). A deep adaptation network study characterized the same problem as distribution shift arising jointly from region and year (Wang et al., 2023), and another single-region crop classifier research reported that a model trained on July imagery misclassified the same crops on May imagery (Sun et al., 2020; Kolluru et al., 2025), which is the within-region analog of the off-season collapse observed in the study. The distinction matters and carries a direct operational consequence because the window is user-controllable, whereas geographic shift is not. Annual-embedding models that dominate the current agricultural foundation model literature cannot measure this effect because they summarize a calendar year and cannot represent within-season timing (Ma et al., 2026; Fang et al., 2026). However, Prithvi's three-frame design makes the phenological window effect measurable. A substantial part of the multi-continent accuracy gradient is ultimately attributable to recoverable temporal misalignment rather than to geographic transfer failure.

### 4.2 Silent confident failure and the limits of deterministic confidence

Off-season deployment reduced OA without a corresponding reduction in confidence, which is the most consequential result for practice. OA fell by up to 0.49 at Kearney and 0.66 at Urbana, while mean confidence remained near its in-season level. ECE, therefore, rose at seven of eight matched pairs, from 0.004 to 0.401 at Kearney and from 0.027 to 0.463 at Urbana (Table 2; Fig. 6). The released checkpoint reported high confidence for these misclassified regions, providing no indication of reduced reliability to an operational user, and high-confidence errors occupied 12 to 51% of the affected scenes. This failure mode is consistent with prior reliability assessments of GeoFMs (Gonzalez-Calabuig et al., 2026), which reported that single-model confidence is insufficient to flag out-of-regime inputs and that ensemble or epistemic uncertainty is required.

A single-model softmax cannot represent epistemic uncertainty, which is the model's uncertainty about whether an input lies within a learned regime. Confidence, therefore, remains high under out-of-regime conditions, where predictions are least reliable (Guo et al., 2017). Predictive entropy from Monte-Carlo head dropout, which carries an epistemic signal (Gal and Ghahramani, 2016), increased off-season at all eight pairs, including Thizy, where expected calibration error was essentially unchanged (0.032 to 0.028). However, predictive entropy still increased from 1.14 to 1.32. These results indicate that the model retains an internal uncertainty signal at the decision layer, but this signal is not expressed in the deterministic confidence value that an operational user reads. This dissociation has been reported in prior crop-mapping work (Mohammadi et al., 2024) and identified as a prerequisite for operational trust (Nejadshamsi et al., 2026; Zhou et al., 2026). Deployment of a released crop model should therefore replace deterministic confidence scores with ensemble or Monte Carlo uncertainty estimates where seasonal misalignment is possible.

### 4.3 Crops, land cover, class composition, and reference nomenclature

No single summary statistic adequately captures the deployed model's performance. Three separable factors govern its apparent accuracy, and each can invert the ranking suggested by any one metric. First, crop and land-cover classes diverge systematically under seasonal misalignment. Crop-class F1 retained only 12–19% of its in-season value off-season (mean change −0.33), whereas land-cover classes retained 66–93% (mean change −0.10), because crop discrimination depends on multi-temporal phenological trajectories that seasonal misalignment removes. In contrast, land-cover classes remain distinguishable by single-date spectral contrast.

Second, class composition governs the divergence between aggregate and class-balanced accuracy. AAFC returned the lowest macro-F1 (0.162) and κ (0.175) despite a mid-range OA (0.457), whereas EUCROPMAP recovered on both metrics (0.245 and 0.306) at the lowest OA (0.402; Fig. 3). This pattern is characteristic of dominant-class-heavy scenes, in which the model attains moderate OA on the majority class while failing on minority classes, thereby lowering class-balanced metrics that weight all classes equally. The same pattern appears in the transfer literature. A random-forest baseline achieved high OA but low macro-F1 on a Chinese dataset in which two classes accounted for 88% of the samples (Wang et al., 2022), and Mohammadi et al. (2024) found that the gains of the adaptation method were concentrated in the minority classes rather than the dominant one. Because the ranking of two deployments can invert depending on whether dominant-class agreement or per-class skill is measured, OA and macro-F1 are informative only when reported together.

Third, reference nomenclature matters because national and continental products define and aggregate crops differently. Coarsening the native 13-class legend to 10 classes improved OA across all 37 events (mean +8.4 percentage points), with the largest gains accruing where the model was furthest from the training distribution (Fig. 8). This reorganization is the deployment-oriented analog of the taxonomic harmonization that benchmark datasets adopt to make multi-country labels comparable (Sykas et al., 2022). The regional pattern indicates that a substantial portion of the OOD shortfall is due to legend mismatch rather than model error. Reference-label quality compounds this effect, because in-season and object-based products report accuracy that varies with training-label quality across crops and regions (Li et al., 2026; Zhou et al., 2026). This indicates that some apparent model error may actually originate in the reference rather than the prediction. Therefore, we adopted a fixed shared label space and an explicit exclude-from-denominator convention to prevent reference-definition effects from being attributed to model behavior.

## 4.4 Operational feasibility and practitioner guidance

The experiments support deployment guidance that requires no retraining. This guidance sets out the conditions under which the released model can and cannot be relied upon. The model can be used with reasonable confidence when: (1) the three-frame window is aligned with the local growing calendar and spans roughly 45–90 days (peaking near 75 days for corn/soybean, 90 for wheat and diverse systems, and 60 for cotton/sorghum); (2) the event is near the training domain; and (3) the legend is coarsened to the 10-class scheme. Window placement is the most consequential and user-controllable decision. A near-real-time window shorter than 30 days costs about 0.11 in OA, and a 120-day arm drifts off-season.

The model should not be relied upon (i) for off-season deployment, where accuracy collapses while deterministic confidence remains high, (ii) for minority-crop mapping in out-of-distribution regions, and (iii) when using deterministic confidence as a quality signal. Fine-tuning or adaptation remains necessary when deployment is far out-of-distribution and a strong class-scheme mismatch persists after coarsening. This requirement is consistent with the surrounding literature, which shows that every successful GeoFM deployment for crop mapping relied on target-region adaptation (Tomotaki-Dawoud et al., 2025; Nejadshamsi et al., 2026; Butsko et al., 2025). Transfer-learning reviews reach the same operating point and recommend unsupervised adaptation for minor to moderate shifts and fine-tuning for large shifts, with 75-85% overall accuracy as the acceptable range for scalable mapping (Long et al., 2025). The near-real-time constraint is consistent with the in-season timing reported by reference producers, for whom discrimination improves as the season progresses and a trade-off exists between accuracy and timely delivery (d'Andrimont et al., 2021; Li et al., 2026). A modern GeoFM, deployed without adaptation, therefore does not remove the alignment step that the pre-foundation-model literature has quantified (Xu et al., 2020; Nyborg et al., 2022; Wang et al., 2022, 2023; Mohammadi et al., 2024). It relocates that step from the training stage to the user's choice of observation window.

## 4.5 Backbone scaling and spectral and contamination robustness

The V1-to-V2 comparison shows that backbone capacity and pretraining scale improve OOD accuracy, with gains widening as distance from the training distribution increases. The collapse of V1 under tight temporal configurations, where V2 retained usable accuracy, indicates that scale also improves robustness to degraded input conditions. Nevertheless, the absolute accuracy under strong shift remained well below the in-distribution ceiling even for the larger model. This aligns with broader assessments that scale alone does not confer reliable transfer across diverse Earth observation tasks (Huang et al., 2026; Marsocci et al., 2024). Backbone upgrades reduce the OOD penalty but do not remove the need for deployment-time alignment.

The spectral-dependency results distinguish discriminative from stabilizing bands. The red band carried the most crop-discriminative information, as expected, because red-edge reflectance and its contrast with the near-infrared region form the physiological basis for vegetation indices widely used in crop mapping (Jain et al., 2024). The NIR and SWIR channels, while less discriminative under mean-fill corruption, were critical for model stability when physically implausible values were injected. This suggests that the model relies on these bands for canopy structure and moisture status cues that must remain within a physically valid range for the representation to function. A parallel finding from a fine-tuned Prithvi yield model attributed its predictions primarily to the NIR and SWIR bands (Nejadshamsi et al., 2026), which indicates that classification and regression tasks share an overlapping spectral core even though they weigh individual bands differently. The practical implication is that quality control should prioritize red-band integrity for discrimination and NIR/SWIR plausibility for model stability. The contamination experiment shows that the three-frame composite provides bounded resilience to single-frame cloud or shadow contamination because self-attention across time can suppress isolated noise (Long et al., 2025). This resilience disappears when all three frames are affected. The threshold structure, reliable below 15%, caution between 15% and 25%, unreliable above 25%, is specific to this model and input configuration, but the underlying principle generalizes. Optical time-series models remain fundamentally bounded by atmospheric conditions in ways that synthetic-aperture-radar products are not (d'Andrimont et al., 2021). For operational deployment in cloudy regions, this indicates that temporal compositing mitigates but does not remove the cloud-cover limitation, and contingency planning for all-frame contamination events is necessary regardless of model capacity.

### 4.6 Limitations

Several limitations bound these conclusions. The evaluation used a single inference run per event. The frozen model is nearly deterministic, so the cross-event spread is treated as the source of variation of interest, and the two residual sources of randomness are reported without seed-level error bars per event. The study also evaluates a single foundation model without a controlled comparison to other geospatial foundation models with performance rankings that vary across tasks and protocols (Corley et al., 2026; Marsocci et al., 2024). The temporal and location-aware backbone was deployed without coordinate or time metadata, and its absence is a plausible contributor to the out-of-distribution degradation, as such metadata conditioning has been reported to aid the model (Szwarcman et al., 2025). The vision-transformer backbone's fixed patch down-sampling limits spatial fidelity on fine-grained and boundary-heavy tasks (Li et al., 2025; Sun et al., 2020), which may compound the minority-class and fragmented-field errors reported here. The Monte Carlo uncertainty reflects only head-level dropout and is a lower bound on the model's epistemic uncertainty, because a fuller treatment would require architectural access not exposed by the released checkpoint (Gal and Ghahramani, 2016). None of these limitations alters the central findings of the out-of-distribution accuracy gradient, the silent off-season failure, and the phenological-alignment result, but each bounds what a deployment-only evaluation of a released checkpoint can establish.

## 5. Conclusions

We evaluated the Prithvi-EO-2.0-600M-TL crop classification model across 37 events in 12 countries spanning three continents. We deployed it to regions and seasons beyond its United States fine-tuning data and validated each event against a regional reference product at the map scale. Overall accuracy declined with distance from the model's United States training distribution, but the aggregate gradient did not account for the whole decline. The class-balanced metrics reordered the deployments by class composition, and much of the apparent cross-continent degradation was due to misalignment between the observation window and local crop phenology rather than geographic transfer failure. The principal result for practice is that the model failed silently when seasonal misalignment occurred. Accuracy collapsed while confidence stayed high, calibration degraded, and the model assigned incorrect crop labels to large fractions of the scene at high confidence and near-zero precision. A user reading the model's confidence, therefore, received no indication that the map was wrong. Four further results shape deployment: coarsening the legend to the granularity the model supports raised accuracy by an average of 8.4 percentage points and improved macro-F1 for every event; the larger V2 backbone helped more as the shift grew; the red band and the multi-frame composite were the model's principal spectral and temporal dependencies; and near-real-time deployment carried a measurable accuracy cost. A fine-tuned crop foundation model, therefore, does not transfer without adjustment. Its confidence cannot be treated as a quality signal when misalignment is possible, and aligning the observation window with local phenology is the primary, most user-controllable determinant of deployment accuracy. These recommendations require no retraining. They are directly actionable by users of the released checkpoint, and they translate the sensitivities that recent reviews have described only qualitatively into reproducible map-level measurements for a model already in operational use.

**Data and code availability.** The pipeline and analysis code, together with a small demonstration dataset, are openly available at https://github.com/venkateshkolluru95/prithvi-eo-crop-ood-evaluation. The released model checkpoints are available through the IBM-NASA Prithvi collection on Hugging Face (https://huggingface.co/ibm-nasa-geospatial) and the TerraTorch toolkit (Gomes et al., 2025). Reference products are available from their providers: the USDA NASS Cropland Data Layer, the Agriculture and Agri-Food Canada Annual Crop Inventory, MapBiomas, and EUCROPMAP (JRC).

**Competing interests.** The authors declare no other competing interests

**Acknowledgments:** We thank all institutes, organizations, and developers of the various datasets for making their products freely available.

**Contributions:** Venkatesh Kolluru: Writing – original draft, Visualization, Validation, Software, Methodology, Investigation, Formal analysis, Data curation, Conceptualization. Rajat Shinde:

Data curation, Conceptualization, Writing – review & editing. Abdelhak Marouane: Writing – review & editing. Caden Helbling: Writing – review & editing. Deepak Shah: Writing – review & editing. Othneil Drew: Writing – review & editing. Iksha Gurung: Writing – review & editing, Project administration. Manil Maskey: Supervision, Resources, Project administration, Funding acquisition. Rahul Ramachandran: Writing – review & editing, Supervision, Resources, Project administration, Funding acquisition.